\documentclass{article}

\usepackage[numbers,compress]{natbib}
\usepackage[main,final]{neurips_2026}

\usepackage[utf8]{inputenc} 
\usepackage[T1]{fontenc}    
\usepackage{hyperref}       
\hypersetup{hidelinks,pdftitle={Learning When to Think: Dual-Reference Offline Optimization for Adaptive VLM Reasoning}, pdfauthor={Hongbin Lin, Sizhe Zou, Juangui Xu, Xinyue Xu, Zhuoli Ouyang, Stanislav Kolchin, Chengwei Qin, Zhongxiang Dai, Yao Shu}}
\usepackage{url}            
\usepackage{booktabs}       
\usepackage{amsfonts}       
\usepackage{nicefrac}       
\usepackage{microtype}      
\usepackage{xcolor}         
\usepackage{amsmath} 
\usepackage{amssymb}
\usepackage{bbm}
\usepackage{graphicx}
\usepackage{multirow} 
\usepackage{placeins}
\usepackage{wrapfig}
\usepackage[most]{tcolorbox}

\usepackage{amsthm}
\usepackage{algorithm}
\usepackage{algpseudocode}
\usepackage{float}
\newtheorem{theorem}{Theorem}
\newtheorem{proposition}{Proposition}
\newtheorem{corollary}{Corollary}
\newtheorem{assumption}{Assumption}

\providecommand{\E}{\mathbb{E}}
\providecommand{\KL}{\mathrm{KL}}
\providecommand{\D}{\mathcal{D}}
\providecommand{\Y}{\mathcal{Y}}

\providecommand{\piF}{\pi_{\mathrm{f}}}
\providecommand{\piS}{\pi_{\mathrm{s}}}
\providecommand{\piRef}{\pi_{\mathrm{ref}}}
\providecommand{\piStar}{\pi^\star}

\providecommand{\rhoDR}{\rho}
\providecommand{\muGen}{\mu}

\title{Learning When to Think: Dual-Reference Offline Optimization for Adaptive VLM Reasoning}

\author{%
  \textbf{Hongbin Lin}\textsuperscript{1} \quad
  \textbf{Sizhe Zou}\textsuperscript{1} \quad
  \textbf{Juangui Xu}\textsuperscript{2} \quad
  \textbf{Xinyue Xu}\textsuperscript{3} \\
  \textbf{Zhuoli Ouyang}\textsuperscript{1} \quad
  \textbf{Stanislav Kolchin}\textsuperscript{1} \quad
  \textbf{Chengwei Qin}\textsuperscript{1,4} \\
  \textbf{Zhongxiang Dai}\textsuperscript{5} \quad
  \textbf{Yao Shu}\textsuperscript{1,4}\thanks{Corresponding author.} \\[4pt]
  \textsuperscript{1}The Hong Kong University of Science and Technology (Guangzhou) \\
  \textsuperscript{2}CISPA Helmholtz Center for Information Security \\
  \textsuperscript{3}The Chinese University of Hong Kong \\
  \textsuperscript{4}The Hong Kong University of Science and Technology \\
  \textsuperscript{5}The Chinese University of Hong Kong, Shenzhen
}

\begin{document}

\maketitle

\begin{abstract}
Reasoning traces improve vision-language models on complex multimodal tasks, but forcing every input to follow a long reasoning process leads to overthinking and unnecessary inference cost. 
We study adaptive fast/slow thinking for VLMs, where one model answers directly for simple inputs and reasons step by step when needed. Fast answers and slow traces often come from distinct behavior policies, making a single-reference offline objective misweight one mode. We propose \textsc{Dual-Reference BOLT} (\textsc{DR-BOLT}), which distinguishes the desired fast/slow mixture reference from the sampling distribution and corrects their mismatch in reward-weighted training. Fast/slow partition competition explains routing in the optimal target; an efficient source-wise correction approximates the exact objective when response formats are well separated. Matched Qwen3-VL comparisons use identical candidate banks to isolate the objective from slow-teacher quality. \textsc{DR-BOLT} improves score--length trade-offs over the evaluated offline and online baselines on ScienceQA and three held-out transfer benchmarks, with a \(24.1\%\) cross-benchmark slow ratio and a \(2.90\times\) end-to-end training speedup over reproduced GRPO. These results support adaptive reasoning in the studied two-reference setting, while broader robustness remains open. Our code is available at https://github.com/HongbinLin3589/DR-BOLT.
\end{abstract}

\section{Introduction}
Many multimodal tasks require more than direct visual recognition. Visual mathematical reasoning, scientific question answering, and expert-level multimodal exams often require a VLM to inspect visual details, combine them with textual context, and perform multi-step inference~\citep{lu2022scienceqa,wang2024mathvision,zou2025dynamath,yue2025mmmupro}. Reasoning traces have therefore become an important mechanism for improving model accuracy on complex inputs. Chain-of-thought and related reasoning methods have shown strong gains in language models~\citep{wei2022chainofthought,wang2022selfconsistency,yao2023treeofthoughts,guo2025deepseekr1,li2026faithful}, and recent VLMs extend this idea by training models to produce explicit intermediate reasoning before answering visual problems~\citep{xu2024llavacot,yao2024mulberry,deng2025openvlthinker,zhan2025visionr1,wang2025vlrethinker,yang2025r1onevision}.

However, always using slow reasoning is inefficient. Many practical VLM inputs can be answered by direct recognition, short visual grounding, or simple text-image matching. For such inputs, forcing a long reasoning trace increases response length, latency, memory usage, and serving cost, and prior studies report that unnecessary thinking can hurt simple cases or create avoidable failure modes~\citep{li2025thinknotthink,kumar2025overthink,hassid2025dontoverthink,xu2026concept}. The desired behavior is therefore input-dependent: a VLM should reason carefully when multi-step inference is needed, but answer directly when slow reasoning provides little benefit.

We study \emph{adaptive thinking} for VLMs, where a single model internalizes when to produce a concise answer and when to generate a longer reasoning process. 
Recent methods explore this goal through thinking and non-thinking modes, length-based mode labels, difficulty-aware rewards, thought dropout, and bi-mode optimization~\citep{yang2025qwen3,lin2025dualmindvlm,xiao2025fast,wang2025ton,yang2025r4b}. 
These methods demonstrate that adaptive thinking can improve the accuracy--efficiency trade-off. 
We focus on a complementary offline setting: generate candidate responses in advance, score them by correctness and efficiency, and train a model to prefer the better response family for each input.

The key observation of this work is that offline adaptive thinking is not merely a problem of controlling response length, but a dual-reference optimization problem. 
Fast direct answers and slow reasoning traces are often generated by distinct behavior policies: concise answers may come from an instruct or no-think policy, while reasoning traces may come from a thinking or chain-of-thought policy. 
This matters for offline reward optimization. 
BOLT decouples response generation from optimization and projects a prompt-normalized, reward-tilted reference distribution through weighted next-token training~\citep{dgo2026}.
Yet standard BOLT assumes a single reference policy. 
For adaptive VLM thinking, forcing fast and slow trajectories into one shared reference can misweight one response family and drive the learned model toward always-fast or always-slow behavior.

We propose \textsc{DR-BOLT}, an offline optimization framework for adaptive VLM thinking. 
Instead of treating direct answers and reasoning traces as two styles sampled from one policy, our method models them as samples from two behavior policies and optimizes under a fast/slow mixture reference. 
We distinguish the desired reference mixture from the distribution used to sample candidates; their ratio corrects the offline weights. Partition competition explains the target's fast/slow allocation, and one trained checkpoint produces adaptive responses at inference. Matched internal comparisons share candidate banks, initialization, rewards, and training budget, including mode-conditioned reward-weighted SFT and multi-reference DPO. This isolates the objective's contribution from the stronger slow teacher. We evaluate ScienceQA~\citep{lu2022scienceqa} in distribution and transfer from ViRL39K to MathVision~\citep{wang2024mathvision}, DynaMath~\citep{zou2025dynamath}, and MMMU-Pro~\citep{yue2025mmmupro}. The latter are held-out evaluations; shared task types and incomplete source-level overlap audits preclude a strict OOD claim.

Our contributions are summarized as follows:
\begin{itemize}
    \item We formulate offline adaptive VLM thinking as a \emph{dual-reference optimization} problem, exposing why single-reference BOLT is mismatched for fast/slow trajectories generated by distinct behavior policies.

    \item We derive the behavior-reference correction and partition-based routing interpretation of \textsc{Dual-Reference BOLT}, with an efficient source-wise approximation. Our contribution is the mismatch diagnosis and its practical correction; the mixture-reference optimum follows standard KL-regularized optimization.

    \item Matched baselines, correction ablations, and sensitivity analyses show improved score--length trade-offs on ScienceQA and three held-out transfer benchmarks, while identifying limitations of the curated two-reference setting.
\end{itemize}

\section{Related Work}
\label{sec:related_work}

\subsection{Reasoning-oriented vision-language models}
Large vision-language models have rapidly advanced multimodal understanding through visual instruction tuning and large-scale image-text alignment~\citep{alayrac2022flamingo,li2023blip2,liu2023llava,dai2023instructblip,bai2023qwenvl,bai2025qwen25vl,bai2025qwen3vl}. Beyond direct perception and instruction following, recent work has explored explicit reasoning in VLMs by transferring chain-of-thought style supervision and reinforcement learning from language models to multimodal tasks~\citep{wei2022chainofthought,wang2024self,shao2024deepseekmath,guo2025deepseekr1}. 
Models such as LLaVA-CoT, Mulberry, OpenVLThinker, Vision-R1, VL-Rethinker, and R1-OneVision train VLMs to generate intermediate reasoning traces before producing final answers~\citep{xu2024llavacot,yao2024mulberry,deng2025openvlthinker,zhan2025visionr1,wang2025vlrethinker,yang2025r1onevision}. These methods improve performance on complex visual reasoning tasks, but they typically encourage slow thinking for all inputs, which increases response length and can cause overthinking on simple examples. 
Our work focuses on adaptive reasoning: preserving the benefits of slow thinking while avoiding unnecessary reasoning when direct answering is sufficient.

\subsection{Adaptive fast/slow thinking in VLMs}
A growing line of work studies how VLMs can decide whether to think. DualMindVLM learns fast and slow thinking modes using length-based mode labels and GRPO, enabling automatic switching between concise and reasoning-based responses~\citep{lin2025dualmindvlm}. FAST-GRPO introduces difficulty-aware rewards and adaptive length regularization to improve the accuracy--efficiency trade-off in large VLM reasoning~\citep{xiao2025fast}. 
TON trains VLMs to select whether to reason through thought dropout followed by reinforcement learning~\citep{wang2025ton}. R-4B builds a general-purpose auto-thinking MLLM with bi-mode annealing and bi-mode policy optimization~\citep{yang2025r4b}. These methods show that adaptive thinking can reduce unnecessary computation while maintaining or improving accuracy. 
However, most existing approaches rely on online RL, explicit mode labels, handcrafted difficulty estimates, or bi-mode rollouts. In contrast, we study a fully offline setting where fast and slow candidate responses are generated in advance and optimized with reward-weighted learning~\citep{mu2026self,wu2025thinking}.

\subsection{Offline reward optimization and the dual-reference gap}
Offline alignment replaces online rollouts with supervised objectives over pre-generated responses. DPO optimizes preference pairs through a reference-regularized objective~\citep{rafailov2023direct}; reward-augmented maximum likelihood and advantage-weighted regression connect reward-weighted learning with policy optimization~\citep{norouzi2016raml,peng2019awr,mu2026drift}. BOLT provides a forward-KL projection onto the Boltzmann target of fixed-reference KL-regularized RL with verifiable rewards~\citep{dgo2026}. We use this framework because our scalar rewards retain correctness and length information for every candidate, and the mixture partition exposes fast/slow allocation. DPO remains a valid alternative; Section~\ref{sec:experiments} compares a mixture-reference DPO baseline and mode-conditioned reward-weighted SFT on the same bank. Our contribution concerns the mismatch between the desired fast/slow reference and the candidate-sampling distribution, rather than a new general solution to KL-regularized optimization.

\section{Preliminaries}
\label{sec:prelim}

\paragraph{Adaptive fast/slow thinking.}
Let \(x=(I,Q)\) denote a multimodal input with image \(I\) and text query \(Q\). 
A VLM can answer \(x\) in two modes. 
In the \emph{fast} mode, it produces a concise direct response \(y_{\mathrm{f}}\). 
In the \emph{slow} mode, it produces a reasoning trace followed by a final answer, denoted as \(y_{\mathrm{s}}=(c,a)\), where \(c\) is the trace and \(a\) is the answer. 
We write the mode as \(z\in\{\mathrm{f},\mathrm{s}\}\). 
Adaptive thinking aims to train a single policy \(\pi_\theta(y|x)\) whose generation behavior internalizes when to answer directly and when to reason step by step, without a separately trained test-time router.

\paragraph{Single-reference BOLT.}
We build on KL-regularized RL with verifiable rewards. 
Given a reference policy \(\piRef\), a reward \(r(x,y)\), and a temperature \(\beta>0\), the objective is
\begin{equation}
    J_{\mathrm{RL}}(\pi_\theta)
    =
    \E_x
    \left[
        \E_{y\sim\pi_\theta(\cdot|x)} r(x,y)
        -
        \beta
        \KL
        \left(
            \pi_\theta(\cdot|x)
            \|
            \piRef(\cdot|x)
        \right)
    \right].
    \label{eq:rlvr}
\end{equation}
BOLT shows that this objective can be optimized offline under a fixed reference policy~\citep{dgo2026}. 
For each input \(x\), the optimal policy is a reward-reweighted reference distribution:
\begin{equation}
    \piStar(y|x)
    =
    \frac{
        \piRef(y|x)
        \exp(r(x,y)/\beta)
    }{
        Z(x)
    },
    \quad
    Z(x)
    =
    \E_{y\sim\piRef(\cdot|x)}
    \left[
        \exp(r(x,y)/\beta)
    \right].
    \label{eq:single_ref_optimal}
\end{equation}
Minimizing the forward KL from \(\piStar\) to \(\pi_\theta\) yields the weighted SFT objective
\begin{equation}
    \mathcal{L}_{\mathrm{BOLT}}(\theta)
    =
    -
    \E_x
    \E_{y\sim\piRef(\cdot|x)}
    \left[
        \frac{
            \exp(r(x,y)/\beta)
        }{
            Z(x)
        }
        \log \pi_\theta(y|x)
    \right].
    \label{eq:bolt_single_ref}
\end{equation}
The derivation is given in Appendix~\ref{app:single_ref_bolt}. 
In practice, \(Z(x)\) is estimated with Monte Carlo samples from the fixed reference. 
Thus, BOLT decouples generation from optimization: responses are generated and weighted once, and the policy is trained by weighted supervised fine-tuning.

\paragraph{Why BOLT for adaptive reasoning?}
BOLT identifies a prompt-normalized Boltzmann target rather than arbitrary reward-weighted SFT, retains scalar reward magnitudes, and exposes fast/slow partition masses. Generation, scoring, and weight construction precede weighted next-token optimization, avoiding repeated online rollouts and verifier calls. DPO~\citep{rafailov2023direct} is also a valid offline alternative, but reward-derived preference pairs retain relative order rather than all reward magnitudes. Section~\ref{sec:experiments} compares mixture-reference DPO and mode-conditioned reward-weighted SFT; we do not claim universal superiority over other offline objectives.

\paragraph{Why single-reference BOLT is insufficient.}
The above formulation assumes that all samples come from one reference distribution \(\piRef\). 
Adaptive VLM thinking often violates this assumption because fast direct answers and slow reasoning traces are commonly generated by different behavior policies. 
A slow trace can have very low likelihood under a fast reference, while a concise answer can have low likelihood under a slow reference. 
Single-reference BOLT can therefore misweight one response family and may lead to always-fast or always-slow behavior. 
The next section formalizes this mismatch as a dual-reference optimization problem.

\section{Dual-Reference BOLT}
\label{sec:dr_bolt}

Standard BOLT assumes a single fixed reference policy. 
Adaptive VLM thinking often violates this assumption because fast direct answers and slow reasoning traces are generated by distinct behavior policies. 
We therefore replace the single reference with a fast/slow mixture reference and derive an offline reward-weighted objective under this dual-reference distribution. 
This formulation reveals two useful structures: routing in the optimal target is governed by fast/slow partition competition, and the behavior-reference correction admits a mode-wise approximation under mode-format separation.

\subsection{Dual-reference objective}

We denote the fast direct-answer policy by \(\piF\) and the slow reasoning policy by \(\piS\). 
We define the desired reference mixture as
\begin{equation}
    \rhoDR(y|x)
    =
    \alpha(x)\piF(y|x)
    +
    \bigl(1-\alpha(x)\bigr)\piS(y|x),
    \label{eq:rho_mix}
\end{equation}
where \(\alpha(x)\in(0,1)\) is the prior probability of the fast mode. 
The actual offline generation process may use a different sampling ratio:
\begin{equation}
    \muGen(y|x)
    =
    q(x)\piF(y|x)
    +
    \bigl(1-q(x)\bigr)\piS(y|x),
    \label{eq:mu_mix}
\end{equation}
where \(q(x)\) is the empirical probability of sampling a fast response. 
The distinction between \(\rhoDR\) and \(\muGen\) is important: \(\rhoDR\) is the reference distribution that defines the regularized target, while \(\muGen\) is the distribution that actually produces the offline data. 
Unless otherwise specified, we use \(\alpha(x)=0.5\) and \(q(x)=0.7\), corresponding to a response bank with \(70\%\) fast candidates and \(30\%\) slow candidates. 
This makes the behavior-reference correction active in the main setting: the mode-wise constants are \(\kappa(y_{\mathrm{f}})=0.5/0.7=0.714\) and \(\kappa(y_{\mathrm{s}})=0.5/0.3=1.667\). 
We additionally use the balanced case \(\alpha=q=0.5\) as a sanity check for the correction term.

We use an accuracy-efficiency reward
\begin{equation}
    r(x,y)
    =
    r_{\mathrm{acc}}(x,y)
    +
    \gamma r_{\mathrm{fmt}}(y)
    -
    \lambda \ell(y),
    \label{eq:reward_dr}
\end{equation}
where \(r_{\mathrm{acc}}\) measures answer correctness, \(r_{\mathrm{fmt}}\) encourages valid answer or reasoning format, and \(\ell(y)\) is the normalized output length. 
The coefficient \(\lambda\) controls the trade-off between accuracy and inference cost.

\begin{theorem}[Dual-reference optimal policy]
\label{thm:dr_optimal}
Consider the KL-regularized objective
\begin{equation}
    J_{\mathrm{DR}}(\pi_\theta)
    =
    \E_x
    \left[
    \E_{y\sim\pi_\theta(\cdot|x)} r(x,y)
    -
    \beta
    \KL
    \left(
        \pi_\theta(\cdot|x)
        \|
        \rhoDR(\cdot|x)
    \right)
    \right].
    \label{eq:dr_rl_objective}
\end{equation}
For each fixed input \(x\), the optimal policy is
\begin{equation}
    \piStar_{\mathrm{DR}}(y|x)
    =
    \frac{
        \rhoDR(y|x)\exp(r(x,y)/\beta)
    }{
        Z_\rho(x)
    },
    \label{eq:dr_optimal}
\end{equation}
where
\begin{equation}
    Z_\rho(x)
    =
    \E_{y\sim\rhoDR(\cdot|x)}
    \left[
        \exp(r(x,y)/\beta)
    \right].
    \label{eq:zrho}
\end{equation}
\end{theorem}

The proof follows the standard Lagrange multiplier derivation and is given in Appendix~\ref{app:proof_dr_optimal}. 
Theorem~\ref{thm:dr_optimal} states that the target policy under the dual-reference objective is a reward-reweighted version of the fast/slow mixture. 
Standard BOLT is recovered as a special case when \(\piF=\piS=\piRef\), in which case \(\rhoDR=\muGen=\piRef\).

\subsection{Router-free adaptive thinking}

The dual-reference partition function decomposes into fast and slow components:
\begin{equation}
    Z_\rho(x)
    =
    \alpha(x)Z_{\mathrm{f}}(x)
    +
    \bigl(1-\alpha(x)\bigr)Z_{\mathrm{s}}(x),
    \label{eq:z_decomp}
\end{equation}
where
\begin{equation}
    Z_{\mathrm{f}}(x)
    =
    \E_{y\sim\piF(\cdot|x)}
    \left[
        \exp(r(x,y)/\beta)
    \right],
    \quad
    Z_{\mathrm{s}}(x)
    =
    \E_{y\sim\piS(\cdot|x)}
    \left[
        \exp(r(x,y)/\beta)
    \right].
    \label{eq:zf_zs}
\end{equation}

\begin{corollary}[Latent routing by partition competition]
\label{cor:routing}
The fast-source marginal of the reward-reweighted joint target is
\begin{equation}
    P^\star(z=\mathrm{f}|x)
    =
    \frac{
        \alpha(x)Z_{\mathrm{f}}(x)
    }{
        \alpha(x)Z_{\mathrm{f}}(x)
        +
        \bigl(1-\alpha(x)\bigr)Z_{\mathrm{s}}(x)
    }.
    \label{eq:p_fast}
\end{equation}
Similarly, \(P^\star(z=\mathrm{s}|x)=1-P^\star(z=\mathrm{f}|x)\).
\end{corollary}

The proof is given in Appendix~\ref{app:proof_routing}. Here \(z\) denotes the reference source in the joint mixture. This marginal corresponds to observable fast/slow output frequencies when sources align with response formats; the trained policy's format-based slow ratio is an empirical diagnostic.
Corollary~\ref{cor:routing} shows that the optimal dual-reference target contains an implicit soft routing rule rather than a separately trained classifier. 
For easy inputs, fast and slow responses may both be correct, but fast responses are shorter, making \(Z_{\mathrm{f}}(x)\) dominate. 
For hard inputs, fast responses may fail while slow responses succeed, making \(Z_{\mathrm{s}}(x)\) dominate despite the length penalty. 
Thus, Eq.~\eqref{eq:p_fast} can be interpreted as a reward-normalized routing score induced by the two references.

\subsection{From dual-reference RL to weighted SFT}

Theorem~\ref{thm:dr_optimal} defines the target policy, but direct sampling from \(\piStar_{\mathrm{DR}}\) is intractable. 
Following BOLT, we minimize the forward KL from the optimal policy to the trainable policy.

\begin{proposition}[Dual-reference weighted SFT]
\label{prop:dr_weighted_sft}
Minimizing
\[
    \E_x
    \left[
    \KL
    \left(
        \piStar_{\mathrm{DR}}(\cdot|x)
        \|
        \pi_\theta(\cdot|x)
    \right)
    \right]
\]
is equivalent, up to constants independent of \(\theta\), to minimizing
\begin{equation}
    \mathcal{L}_{\rho}(\theta)
    =
    -
    \E_x
    \E_{y\sim\rhoDR(\cdot|x)}
    \left[
        \frac{
            \exp(r(x,y)/\beta)
        }{
            Z_\rho(x)
        }
        \log \pi_\theta(y|x)
    \right].
    \label{eq:rho_loss}
\end{equation}
If samples are drawn from \(\muGen\) rather than \(\rhoDR\), and \(\rhoDR\) is absolutely continuous with respect to \(\muGen\), the corrected offline objective is
\begin{equation}
    \mathcal{L}_{\mu}(\theta)
    =
    -
    \E_x
    \E_{y\sim\muGen(\cdot|x)}
    \left[
        \kappa(x,y)
        \frac{
            \exp(r(x,y)/\beta)
        }{
            Z_\rho(x)
        }
        \log \pi_\theta(y|x)
    \right],
    \label{eq:mu_loss}
\end{equation}
where
\begin{equation}
    \kappa(x,y)
    =
    \frac{\rhoDR(y|x)}{\muGen(y|x)}.
    \label{eq:correction}
\end{equation}
\end{proposition}

The proof is a forward-KL expansion plus a change of measure; see Appendix~\ref{app:proof_dr_weighted_sft}. 
Proposition~\ref{prop:dr_weighted_sft} shows that the offline weight has two factors: a reward-induced factor and a behavior-reference correction factor. Exact correction remains valid under arbitrary overlap; mode separation is required only for the pointwise mode-wise approximation.
When \(q(x)=\alpha(x)\), we have \(\muGen=\rhoDR\), so \(\kappa(x,y)=1\) exactly. 
This balanced case is a sanity check, while the main non-balanced setting exposes and validates the correction term.

\subsection{Practical mode-wise correction}

Exact computation of \(\kappa(x,y)\) requires evaluating both \(\piF\) and \(\piS\) on every sampled response, which is expensive for large VLMs. 
We use the following empirical structure of fast/slow outputs: the two modes are often separated by response format and generation prefix, while length further amplifies the mismatch for long reasoning traces. 
This motivates an approximate mode-format separation assumption. Offline source labels are recorded from the generating policy, rather than inferred from response length or a format classifier.

\begin{assumption}[Approximate mode-format separation]
\label{assump:sep}
For fast samples \(y_{\mathrm{f}}\) and slow samples \(y_{\mathrm{s}}\), define
\[
    \delta_{\mathrm{f}}(x,y_{\mathrm{f}})
    =
    \frac{\piS(y_{\mathrm{f}}|x)}{\piF(y_{\mathrm{f}}|x)},
    \quad
    \delta_{\mathrm{s}}(x,y_{\mathrm{s}})
    =
    \frac{\piF(y_{\mathrm{s}}|x)}{\piS(y_{\mathrm{s}}|x)}.
\]
We assume there exist small constants \(\varepsilon_{\mathrm{f}},\varepsilon_{\mathrm{s}}\ll 1\) such that
\begin{equation}
    \delta_{\mathrm{f}}(x,y_{\mathrm{f}})\leq \varepsilon_{\mathrm{f}},
    \quad
    \delta_{\mathrm{s}}(x,y_{\mathrm{s}})\leq \varepsilon_{\mathrm{s}}.
    \label{eq:support_sep}
\end{equation}
The constants capture mode-format mismatch. Both references and their offline samples are fixed, so these ratios do not evolve as the trainable policy changes during Phase 2.
In practice, \(\varepsilon_{\mathrm{s}}\) may be further reduced by length compounding on long reasoning traces, while \(\varepsilon_{\mathrm{f}}\) may be dominated by prefix-level mismatch.
\end{assumption}

\begin{proposition}[Mode-wise correction]
\label{prop:modewise}
Under Assumption~\ref{assump:sep}, if \(\alpha(x),q(x)\in[\eta,1-\eta]\) for some \(\eta>0\), then
\begin{equation}
    \kappa(x,y_{\mathrm{f}})
    =
    \frac{\alpha(x)}{q(x)}
    +
    O(\varepsilon_{\mathrm{f}}),
    \quad
    \kappa(x,y_{\mathrm{s}})
    =
    \frac{1-\alpha(x)}{1-q(x)}
    +
    O(\varepsilon_{\mathrm{s}}).
    \label{eq:modewise}
\end{equation}
\end{proposition}

The proof is given in Appendix~\ref{app:proof_modewise}. 
Proposition~\ref{prop:modewise} shows that dual-reference correction can be approximated without expensive dual-model likelihood evaluation when mode-format separation holds. 
Appendix~\ref{app:support_validation} validates the assumption with a cross-likelihood matrix, and Appendix~\ref{app:kappa_validation} compares exact and mode-wise corrections.
The exact residual magnitudes are \(|q-\alpha|\delta_{\mathrm{f}}/[q(q+(1-q)\delta_{\mathrm{f}})]\) and \(|q-\alpha|\delta_{\mathrm{s}}/[(1-q)(1-q+q\delta_{\mathrm{s}})]\), so overlap increases pointwise approximation error smoothly. On the main path, exact and mode-wise log corrections have correlation \(0.9902\), with \(1.2\%\) of samples differing by more than \(0.01\). An ablation path has \(3\)--\(5\%\) reasoning-style completions in the fast pool and lower correlation \(0.9730\). Exact correction is available for ambiguous samples; filtering reasoning markers or fast-pool outputs longer than \(30\) generated tokens is another format-control option.

\paragraph{Single-reference collapse mechanism.}
Assumption~\ref{assump:sep} also explains why single-reference BOLT can collapse in this setting. 
If a fast reference is used for all samples, then slow samples are suppressed by a factor proportional to \(\delta_{\mathrm{s}}\). 
If a slow reference is used for all samples, then fast samples are suppressed by a factor proportional to \(\delta_{\mathrm{f}}\). 
Thus, single-reference BOLT does not merely use a weaker approximation; under mode separation, it systematically downweights one response family and can drive the model toward always-fast or always-slow behavior.

\subsection{Empirical objective}

For each input \(x\), we sample fast and slow response sets:
\[
    \Y_{\mathrm{f}}(x)=\{y_i^{\mathrm{f}}\}_{i=1}^{N_{\mathrm{f}}},
    \quad
    y_i^{\mathrm{f}}\sim\piF(\cdot|x),
\]
\[
    \Y_{\mathrm{s}}(x)=\{y_j^{\mathrm{s}}\}_{j=1}^{N_{\mathrm{s}}},
    \quad
    y_j^{\mathrm{s}}\sim\piS(\cdot|x).
\]
We estimate
\begin{equation}
    \widehat{Z}_{\mathrm{f}}(x)
    =
    \frac{1}{N_{\mathrm{f}}}
    \sum_{i=1}^{N_{\mathrm{f}}}
    \exp(r(x,y_i^{\mathrm{f}})/\beta),
    \quad
    \widehat{Z}_{\mathrm{s}}(x)
    =
    \frac{1}{N_{\mathrm{s}}}
    \sum_{j=1}^{N_{\mathrm{s}}}
    \exp(r(x,y_j^{\mathrm{s}})/\beta),
    \label{eq:z_hat_fs}
\end{equation}
and
\begin{equation}
    \widehat{Z}_{\rho}(x)
    =
    \alpha(x)\widehat{Z}_{\mathrm{f}}(x)
    +
    \bigl(1-\alpha(x)\bigr)\widehat{Z}_{\mathrm{s}}(x).
    \label{eq:z_hat_rho}
\end{equation}

The practical DR-BOLT weight is
\begin{equation}
    w_{\mathrm{DR}}(x,y)
    =
    \widehat{\kappa}(x,y)
    \cdot
    \frac{
        \exp(r(x,y)/\beta)
    }{
        \widehat{Z}_{\rho}(x)
    },
    \label{eq:dr_weight}
\end{equation}
where \(\widehat{\kappa}\) is either the exact correction in Eq.~\eqref{eq:correction} or the mode-wise approximation in Eq.~\eqref{eq:modewise}. 
The final training objective is
\begin{equation}
    \boxed{
    \mathcal{L}_{\mathrm{DR\text{-}BOLT}}(\theta)
    =
    -
    \sum_{(x,y)\in\D_{\mathrm{DR}}}
    w_{\mathrm{DR}}(x,y)
    \log \pi_\theta(y|x)
    }.
    \label{eq:dr_bolt_loss}
\end{equation}

\subsection{Training algorithm}

DR-BOLT uses three phases. 
Phase 0 performs a short mixed warm-up, because the mixture reference \(\rhoDR\) is a distribution rather than a single physical checkpoint. 
This warm-up is used to initialize the trainable policy near both response formats before reward-weighted optimization. 
Phase 1 generates fast and slow responses and computes DR-BOLT weights. 
Phase 2 optimizes weighted next-token likelihood.
The references, response bank, rewards, mixture parameters, and correction weights are frozen before Phase 2; no on-policy rollouts recompute these weights.

\begin{algorithm}[t]
\centering
\footnotesize
\caption{\textsc{Dual-Reference BOLT}}
\label{alg:dr_bolt}
\begin{algorithmic}[1]
\Require Inputs \(\mathcal{X}\), fast policy \(\piF\), slow policy \(\piS\), prior \(\alpha(x)\), sampling ratio \(q(x)\), samples \(N_{\mathrm{f}},N_{\mathrm{s}}\), reward \(r\), temperature \(\beta\)
\Statex \textbf{Phase 0: Mixed warm-up}
\State Train \(\pi_\theta\) on a small unweighted mixture of fast and slow samples.
\Statex \textbf{Phase 1: Dual-reference generation}
\State Initialize \(\D_{\mathrm{DR}}\leftarrow\emptyset\).
\For{each \(x\in\mathcal{X}\)}
    \State Sample \(\Y_{\mathrm{f}}(x)=\{y_i^{\mathrm{f}}\}_{i=1}^{N_{\mathrm{f}}}\) from \(\piF(\cdot|x)\).
    \State Sample \(\Y_{\mathrm{s}}(x)=\{y_j^{\mathrm{s}}\}_{j=1}^{N_{\mathrm{s}}}\) from \(\piS(\cdot|x)\).
    \State Compute \(r(x,y)\) for all \(y\in\Y_{\mathrm{f}}(x)\cup\Y_{\mathrm{s}}(x)\).
    \State Estimate \(\widehat{Z}_{\mathrm{f}}(x)\), \(\widehat{Z}_{\mathrm{s}}(x)\), and \(\widehat{Z}_{\rho}(x)\).
    \State Compute \(w_{\mathrm{DR}}(x,y)\) for all sampled responses using Eq.~\eqref{eq:dr_weight}.
    \State Add all weighted triples \((x,y,w_{\mathrm{DR}}(x,y))\) to \(\D_{\mathrm{DR}}\).
\EndFor
\Statex \textbf{Phase 2: Offline optimization}
\State Optimize \(\pi_\theta\) on \(\D_{\mathrm{DR}}\) using Eq.~\eqref{eq:dr_bolt_loss}.
\State \Return adaptive policy \(\pi_\theta\).
\end{algorithmic}
\end{algorithm}
\vspace{-15pt}

\subsection{Discussion}
\label{sec:dr_bolt_discussion}

DR-BOLT differs from standard BOLT by changing the reference distribution from a single policy to a fast/slow mixture. 
This makes the optimal policy, partition function, routing probability, and sample weights mode-dependent. 
The formulation provides a router-free mechanism in the target distribution: fast responses dominate when their reward-weighted partition mass is larger, while slow responses dominate when their correctness gain outweighs their length cost. 
The behavior-reference correction \(\kappa\) is needed because fast and slow trajectories come from different policies, and Proposition~\ref{prop:modewise} makes this correction practical under mode-format separation. 
When the two policies produce highly overlapping formats, exact correction or additional filtering may be needed; we empirically examine routing calibration, mode collapse, mode-format separation, and exact-versus-mode-wise correction in Section~\ref{sec:experiments}.
Both behaviors are useful when each has a reward advantage on part of the input distribution. If slow reasoning adds no compensating correctness benefit, concentration toward fast responses is the intended reward-optimal solution, rather than pathological collapse. A reference's single-decoding accuracy is not a theoretical ceiling because reward tilting and multi-candidate coverage can improve the target distribution. Nevertheless, the population target remains on the mixture support, and the empirical method receives reward evidence only from sampled candidates; it provides no unconstrained online exploration.

\section{Experiments}
\label{sec:experiments}

\subsection{Setup}

We train on the ScienceQA training split and evaluate on its official test split~\citep{lu2022scienceqa}. We also study cross-benchmark transfer by training on ViRL39K~\citep{wang2025vlrethinker} and evaluating on MathVision~\citep{wang2024mathvision}, DynaMath~\citep{zou2025dynamath}, and MMMU-Pro~\citep{yue2025mmmupro}. These are held-out evaluations rather than strictly verified source- or template-level OOD settings. We add a second same-benchmark evaluation on a constructed disjoint 3DSRBench split~\citep{ma20253dsrbench} (Appendix~\ref{app:3dsrbench}).

Our fast and slow references are Qwen3-VL-8B-Instruct and Qwen3-VL-8B-Thinking. Matched offline methods share the candidate bank, reward, initialization, and optimization budget. In addition to the baselines in Table~\ref{tab:main_internal}, we evaluate multi-reference DPO (MR-DPO) and mode-conditioned reward-weighted SFT (Cond-RW). We reproduce GRPO, FAST-GRPO~\citep{xiao2025fast}, TON~\citep{wang2025ton}, and DualMindVLM~\citep{lin2025dualmindvlm} using the common training inputs and evaluation pipeline; their method-specific rewards remain different. Appendix~\ref{app:exp_details} and Appendix~\ref{app:rebuttal_experiments} give protocols and implementation details.

Length is mean response word count, excluding the prompt. Slow ratio is format-based, using the reasoning marker or slow-thinking format rather than a length threshold. In mixed benchmark tables, this ratio aggregates only MathVision, DynaMath, and MMMU-Pro evaluations of the ViRL39K-trained checkpoint; ScienceQA uses a separately trained checkpoint.

\subsection{Main Results}

Tables~\ref{tab:main_external} and~\ref{tab:main_internal} report \(\text{Score}/\text{Length}\). Scores use the same benchmark-specific answer extraction across methods. Training time is measured end to end on four A100-80GB GPUs: \textsc{DR-BOLT} includes offline fast/slow candidate generation, reward computation, weight construction, and policy optimization; online baselines include rollout generation and updates. Memory is peak per-GPU usage, and speedup is relative to reproduced GRPO.

\begin{table*}[t]
\centering
\small
\caption{
Our reproductions of adaptive-thinking methods and GRPO. ScienceQA uses same-benchmark training; the other datasets measure transfer after ViRL39K training. Cells report \(\text{Score}/\text{Length}\) in words. Runtime includes candidate generation or online rollouts and all training stages; memory is peak per GPU. Speedup is relative to GRPO.
}
\label{tab:main_external}
\resizebox{\textwidth}{!}{
\begin{tabular}{lccccc}
\toprule
\multirow{2}{*}{Dataset}
& GRPO
& FAST-GRPO~\citep{xiao2025fast}
& TON~\citep{wang2025ton}
& DualMindVLM~\citep{lin2025dualmindvlm}
& \textbf{DR-BOLT (Ours)} \\
& \multicolumn{5}{c}{\textbf{Score / Length}} \\
\midrule
ScienceQA (IID)
& \(0.962 / 201.0\)
& \(0.964 / 37.6\)
& \(0.963 / 35.9\)
& \(0.872 / 87.2\)
& \(\mathbf{0.967 / 20.4}\) \\
MathVision (transfer)
& \(0.463 / 2498.2\)
& \(0.527 / 1245.7\)
& \(0.488 / 1505.8\)
& \(0.302 / 446.0\)
& \(\mathbf{0.606 / 1058.1}\) \\
DynaMath (transfer)
& \(0.769 / 1402.2\)
& \(0.782 / 512.4\)
& \(0.775 / 495.5\)
& --
& \(\mathbf{0.799 / 370.6}\) \\
MMMU-Pro (transfer)
& \(0.498 / 2006.4\)
& \(0.540 / 892.6\)
& \(0.519 / 747.3\)
& --
& \(\mathbf{0.581 / 385.7}\) \\
\midrule
Training Type
& online RL / GRPO
& online RL / GRPO
& online RL / GRPO
& online RL / GRPO
& offline dual-ref BOLT \\
Time (h)
& \(202.4\)
& \(176.6\)
& \(248.9\)
& --
& \(\mathbf{69.7}\) \\
Memory (GB)
& \(70.2\)
& \(71.8\)
& \(67.5\)
& \(66.2\)
& \(\mathbf{60.2}\) \\
Speedup
& \(1.00\times\)
& \(1.15\times\)
& \(0.81\times\)
& --
& \(\mathbf{2.90\times}\) \\
\bottomrule
\end{tabular}
}
\vspace{-5pt}
\end{table*}

\begin{table*}[t]
\centering
\small
\caption{
Matched internal comparisons. Cells report \(\text{Score}/\text{Length}\) in words; ScienceQA and cross-benchmark results use separately trained checkpoints. Slow ratio (\%) aggregates only MathVision, DynaMath, and MMMU-Pro.
}
\label{tab:main_internal}
\resizebox{\textwidth}{!}{
\begin{tabular}{lccccccc}
\toprule
\multirow{2}{*}{Dataset}
& Qwen3-Instruct
& Qwen3-Thinking
& Mix SFT
& Reward SFT
& Single-ref Fast
& Single-ref Slow
& \textbf{DR-BOLT} \\
& \multicolumn{7}{c}{\textbf{Score / Length}} \\
\midrule
ScienceQA (IID)
& \(0.943 / 7.2\)
& \(0.954 / 288.3\)
& \(0.955 / 100.1\)
& \(0.956 / 89.3\)
& \(0.945 / 9.3\)
& \(0.955 / 317.7\)
& \(\mathbf{0.967 / 20.4}\) \\
MathVision (transfer)
& \(0.244 / 395.8\)
& \(0.368 / 5775.3\)
& \(0.406 / 3777.6\)
& \(0.421 / 3802.9\)
& \(0.301 / 474.7\)
& \(0.439 / 5901.8\)
& \(\mathbf{0.606 / 1058.1}\) \\
DynaMath (transfer)
& \(0.613 / 205.5\)
& \(0.654 / 2696.7\)
& \(0.730 / 1896.1\)
& \(0.741 / 1675.0\)
& \(0.627 / 241.4\)
& \(0.752 / 2022.9\)
& \(\mathbf{0.799 / 370.6}\) \\
MMMU-Pro (transfer)
& \(0.278 / 265.3\)
& \(0.377 / 3082.3\)
& \(0.472 / 1352.6\)
& \(0.484 / 1105.9\)
& \(0.412 / 443.3\)
& \(0.408 / 2575.1\)
& \(\mathbf{0.581 / 385.7}\) \\
\midrule
Slow Ratio (\%)
& \(0.0\)
& \(99.8\)
& \(67.2\)
& \(60.4\)
& \(1.9\)
& \(97.4\)
& \(24.1\)\\
\bottomrule
\end{tabular}
}
\vspace{-5pt}
\end{table*}

\begin{figure}[t]
\centering
\includegraphics[width=0.90\linewidth]{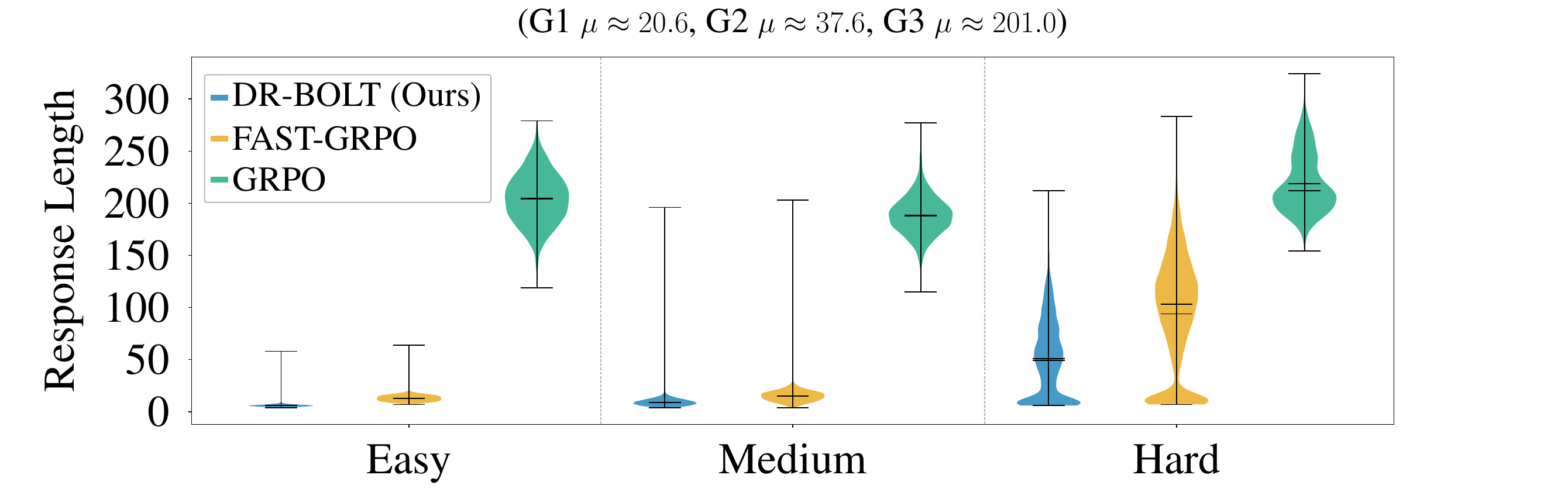}
\vspace{-5pt}
\caption{
Difficulty-wise response-Length distributions. 
Each group contains the response-Length distributions for \textsc{DR-BOLT}, FAST-GRPO, and GRPO. 
\textsc{DR-BOLT} stays short on easy examples and increases reasoning Length on harder examples, while GRPO-style training remains much longer.
\vspace{-10pt}
}
\label{fig:difficulty_routing}
\end{figure}

\paragraph{Comparison with external adaptive-thinking baselines.}
\textsc{DR-BOLT} has the highest score among the evaluated methods on all four benchmarks. Relative to reproduced GRPO, its score gains are \(0.005\), \(0.143\), \(0.030\), and \(0.083\), and responses are \(9.85\times\), \(2.36\times\), \(3.78\times\), and \(5.20\times\) shorter, respectively. End-to-end training takes \(69.7\) rather than \(202.4\) hours (\(2.90\times\) speedup), with peak per-GPU memory of \(60.2\) rather than \(70.2\)GB. These are method-level comparisons under our common-data reproduction protocol.

\paragraph{Internal baselines and single-reference collapse.}
The matched response bank controls for slow-teacher quality: all offline methods see the same fast and slow candidates. Mix SFT and Reward SFT remain longer and lower-scoring than \textsc{DR-BOLT}. Single-ref Fast and Single-ref Slow have cross-benchmark slow ratios of \(1.9\%\) and \(97.4\%\), versus \(24.1\%\) for \textsc{DR-BOLT}, consistent with systematic mode suppression in these runs.

\paragraph{Stronger offline alternatives and sensitivity.}
MR-DPO uses the same mixture reference and reward-ranked preference pairs; Cond-RW explicitly learns a fast/slow source token. Under matched candidate banks and response-token budgets, their MathVision results are \(0.568/1398.4\) and \(0.513/1810.7\), versus \(0.606/1058.1\) for \textsc{DR-BOLT}. The same ranking holds on the other three benchmarks (Table~\ref{tab:offline_alternatives}). This supports the observed advantage for this scalar-reward bank, without establishing universal superiority over preference optimization. One-factor sensitivity studies show that \(\alpha\) controls slow usage, while changing \(q\) from \(0.7\) to \(0.3\) changes ScienceQA score by \(0.002\) after correction. Increasing \(\lambda\) from \(0\) to \(0.10\) reduces ScienceQA Length from \(28.9\) to \(15.7\) words; scores lie in \([0.964,0.969]\). The shared main setting is not tuned separately on the transfer test sets (Appendix~\ref{app:sensitivity}).

\subsection{Score--Length Dynamics}

\begin{wrapfigure}{r}{0.48\linewidth}
\vspace{-15pt}
\centering
\includegraphics[width=\linewidth]{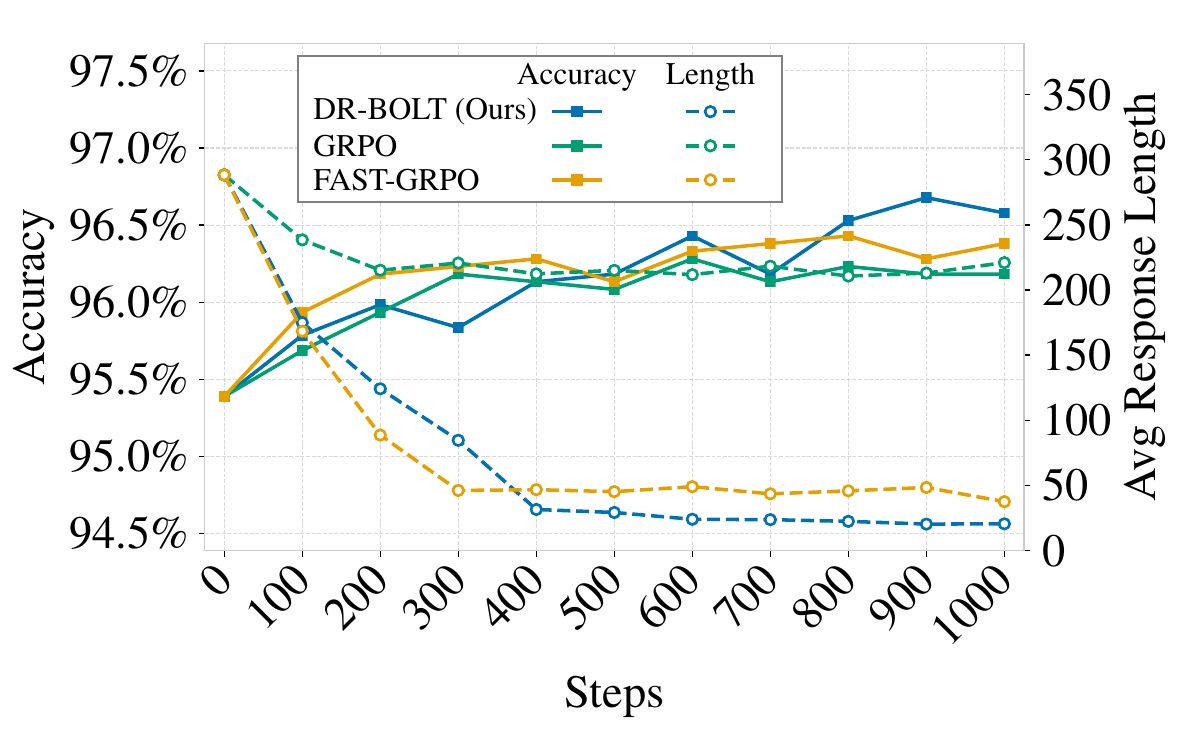}
\vspace{-15pt}
\caption{
Score--Length dynamics on ScienceQA. 
Solid lines show accuracy and dashed lines show response Length.
}
\label{fig:pareto}
\vspace{-15pt}
\end{wrapfigure}

Figure~\ref{fig:pareto} tracks the training dynamics on ScienceQA. 
\textsc{DR-BOLT} rapidly moves toward the upper-left region of the score--Length space: it increases accuracy from the initial slow model while reducing Length from \(288.3\) to about \(20.4\). 
GRPO improves accuracy but remains much longer, staying above \(200\) words at the end of training. 
FAST-GRPO reduces Length more aggressively than GRPO, but converges to a lower final score and longer responses than \textsc{DR-BOLT}. 
This result illustrates the benefit of offline reward-weighted optimization under a dual-reference mixture: the model learns to keep the accuracy benefit of reasoning while dropping unnecessary reasoning tokens on easy or already-solved examples.

\subsection{Routing Behavior}

We next evaluate whether \textsc{DR-BOLT} learns difficulty-dependent routing rather than simply shortening all responses. 
Figure~\ref{fig:difficulty_routing} shows response-Length distributions across easy, medium, and hard groups. 
The desired behavior is not a fixed response length: easy examples should be answered directly, while harder examples should activate longer reasoning more often. 
\textsc{DR-BOLT} follows this pattern, producing very short responses on easy examples and increasing Length on harder examples. 
By contrast, GRPO-style training remains substantially longer across difficulty groups, reflecting less selective use of slow reasoning. 
This supports the claim that \textsc{DR-BOLT} learns input-dependent allocation of reasoning effort.

\subsection{Partition-based Routing Calibration}

\begin{wrapfigure}{r}{0.5\linewidth}
\centering
\includegraphics[width=\linewidth]{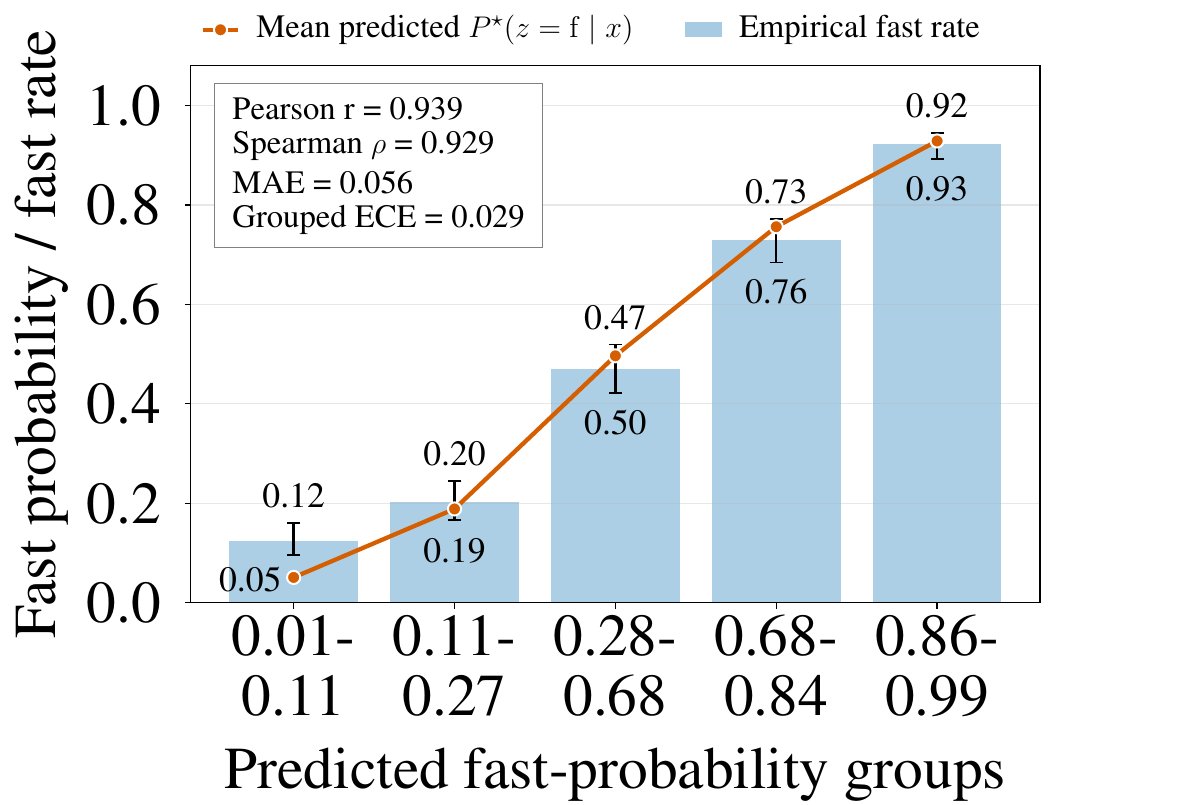}
\vspace{-5pt}
\caption{
Partition-based routing calibration.
Bars show empirical fast rate and markers show predicted fast probability.
}
\vspace{-5pt}
\label{fig:partition_routing}
\end{wrapfigure}

Finally, we validate the partition-competition mechanism in Corollary~\ref{cor:routing}. 
For each input, we estimate \(\widehat{Z}_{\mathrm{f}}(x)\) and \(\widehat{Z}_{\mathrm{s}}(x)\), compute the predicted fast probability
\[
P^\star(z=\mathrm{f}|x)
=
\frac{\alpha(x)\widehat{Z}_{\mathrm{f}}(x)}
{\alpha(x)\widehat{Z}_{\mathrm{f}}(x)+(1-\alpha(x))\widehat{Z}_{\mathrm{s}}(x)},
\]
and compare it with the empirical fast rate under repeated auto-mode decoding. 
Because each prompt is decoded only a finite number of times, we group prompts into equal-frequency bins according to predicted fast probability. 
Figure~\ref{fig:partition_routing} shows that the empirical fast rate closely follows the predicted fast probability across groups. 
Across 100 prompts, the predicted probability and empirical fast rate have Pearson correlation \(r=0.939\), Spearman correlation \(\rho=0.929\), and grouped calibration error \(0.029\). 
These results indicate that the model's routing behavior is well explained by fast/slow partition competition rather than by an external classifier or random length variation.

\subsection{Controlled Offline Comparisons}
\label{sec:controlled_offline}

\begin{table}[H]
\centering
\small
\caption{Controlled offline alternatives on identical candidate banks. Cells report score / response words. ScienceQA uses a separate checkpoint; slow ratio aggregates the three transfer benchmarks.}
\label{tab:offline_alternatives}
\resizebox{0.8\linewidth}{!}{%
\begin{tabular}{lccccc}
\toprule
Method & ScienceQA & MathVision & DynaMath & MMMU-Pro & Slow ratio (\%) \\
\midrule
MR-DPO & \(0.964/31.8\) & \(0.568/1398.4\) & \(0.787/506.9\) & \(0.557/528.6\) & \(29.4\) \\
Cond-RW & \(0.960/48.6\) & \(0.513/1810.7\) & \(0.761/714.8\) & \(0.521/689.5\) & \(39.7\) \\
\textsc{DR-BOLT} & \(0.967/20.4\) & \(0.606/1058.1\) & \(0.799/370.6\) & \(0.581/385.7\) & \(24.1\) \\
\bottomrule
\end{tabular}%
}
\vspace{-10pt}
\end{table}

MR-DPO and Cond-RW share the same candidate bank, reward, frozen references, initialization, and optimization budget as \textsc{DR-BOLT}. Table~\ref{tab:offline_alternatives} isolates the objective's gain from the stronger slow teacher: \textsc{DR-BOLT} improves both score and length on every benchmark in this comparison.

\paragraph{Ablation of behavior-reference correction.}
With the ViRL39K bank and optimization setup fixed at \(\alpha=0.5,q=0.7\), removing \(\kappa\) drives the cross-benchmark slow ratio to \(0.13\%\) and scores to \(0.260/0.626/0.314\) on MathVision/DynaMath/MMMU-Pro. Exact correction restores a \(22.0\%\) slow ratio and scores of \(0.623/0.804/0.594\); mode-wise correction yields \(24.1\%\) and \(0.606/0.799/0.581\). These controls attribute the restored trade-off to the correction applied to identical candidates (Appendix~\ref{app:kappa_downstream}).

Exact and mode-wise correction differ by at most \(0.017\) in score in these runs.

\section{Conclusion}
\label{sec:conclusion}

We presented \textsc{DR-BOLT} for offline adaptive VLM reasoning. Distinguishing the desired fast/slow reference from the candidate-sampling distribution yields a behavior-reference correction and a partition-based interpretation of target routing. Controlled comparisons with identical candidate banks show improved score--length trade-offs over the evaluated offline alternatives on ScienceQA and held-out transfer benchmarks. End-to-end accounting includes candidate generation and shows lower training cost than reproduced GRPO. Future work may extend the formulation to more reasoning modes, continuous budgets, bank refresh, and offline-online multimodal reasoning.

\clearpage
\section{Acknowledgement}
This work was supported by the Guangdong Basic and Applied Basic Research Foundation (Grant No. 2025A1515110115).
\bibliography{cite}
\bibliographystyle{abbrvnat}

\clearpage
\appendix

\section{Notation}
\label{app:notation}

\begin{table}[h]
\centering
\small
\caption{Notation used in the paper.}
\label{tab:notation}
\begin{tabular}{lp{0.72\linewidth}}
\toprule
Symbol & Meaning \\
\midrule
\(x=(I,Q)\) & Multimodal input consisting of image \(I\) and text query \(Q\). \\
\(y\) & Model response. \\
\(y_{\mathrm{f}}\) & Fast response, usually a concise direct answer. \\
\(y_{\mathrm{s}}=(c,a)\) & Slow response containing reasoning trace \(c\) and final answer \(a\). \\
\(z\in\{\mathrm{f},\mathrm{s}\}\) & Latent response mode: fast or slow. \\
\(\pi_\theta(y|x)\) & Trainable VLM policy. \\
\(\piF(y|x)\) & Fast behavior/reference policy. \\
\(\piS(y|x)\) & Slow behavior/reference policy. \\
\(\piRef(y|x)\) & Single reference policy used in standard BOLT. \\
\(\piStar(y|x)\) & Optimal policy under single-reference BOLT. \\
\(\piStar_{\mathrm{DR}}(y|x)\) & Optimal policy under dual-reference BOLT. \\
\(\rhoDR(y|x)\) & Desired dual-reference mixture distribution. \\
\(\muGen(y|x)\) & Empirical data-generating mixture distribution. \\
\(\alpha(x)\) & Desired fast-mode prior in \(\rhoDR\). \\
\(q(x)\) & Empirical fast sampling ratio in \(\muGen\). \\
\(r(x,y)\) & Scalar reward for response \(y\) on input \(x\). \\
\(r_{\mathrm{acc}}(x,y)\) & Correctness reward. \\
\(r_{\mathrm{fmt}}(y)\) & Format reward. \\
\(\ell(y)\) & Normalized length or compute cost used in the reward. \\
\(\gamma,\lambda\) & Weights for format reward and length penalty. \\
\(\beta\) & Temperature in KL-regularized RL / BOLT weighting. \\
\(Z(x)\) & Single-reference partition function. \\
\(Z_\rho(x)\) & Dual-reference partition function. \\
\(Z_{\mathrm{f}}(x), Z_{\mathrm{s}}(x)\) & Fast and slow reward-weighted partition functions. \\
\(\widehat Z_{\mathrm{f}}(x), \widehat Z_{\mathrm{s}}(x), \widehat Z_\rho(x)\) & Monte Carlo estimates of the corresponding partition functions. \\
\(P^\star(z=\mathrm{f}|x)\) & Optimal fast-mode marginal probability induced by partition competition. \\
\(\kappa(x,y)\) & Exact behavior-reference correction \(\rhoDR(y|x)/\muGen(y|x)\). \\
\(\widehat\kappa(x,y)\) & Practical correction used in empirical training. \\
\(\delta_{\mathrm{f}}(x,y_{\mathrm{f}})\) & Cross-mode likelihood ratio \(\piS(y_{\mathrm{f}}|x)/\piF(y_{\mathrm{f}}|x)\). \\
\(\delta_{\mathrm{s}}(x,y_{\mathrm{s}})\) & Cross-mode likelihood ratio \(\piF(y_{\mathrm{s}}|x)/\piS(y_{\mathrm{s}}|x)\). \\
\(\varepsilon_{\mathrm{f}},\varepsilon_{\mathrm{s}}\) & Small constants controlling approximate mode-format separation. \\
\(\eta\) & Lower bound ensuring \(\alpha(x),q(x)\in[\eta,1-\eta]\). \\
\(\Y_{\mathrm{f}}(x),\Y_{\mathrm{s}}(x)\) & Sampled fast and slow candidate response sets for input \(x\). \\
\(N_{\mathrm{f}},N_{\mathrm{s}}\) & Number of fast and slow samples per input. \\
\(\D_{\mathrm{DR}}\) & Weighted dual-reference training dataset. \\
\(w_{\mathrm{DR}}(x,y)\) & Final DR-BOLT sample weight. \\
\(\mathcal{L}_{\mathrm{BOLT}}\) & Single-reference BOLT weighted SFT objective. \\
\(\mathcal{L}_{\mathrm{DR\text{-}BOLT}}\) & Dual-reference BOLT weighted SFT objective. \\
\(\E[\cdot]\) & Expectation. \\
\(\KL(\cdot\|\cdot)\) & Kullback--Leibler divergence. \\
Length & Response length reported in experiments, measured by word count. \\
Slow Ratio & Fraction of generated responses classified as slow-thinking responses. \\
\bottomrule
\end{tabular}
\end{table}

\section{Limitations}
\label{sec:limitations}

Our study focuses on adaptive fast/slow thinking with two response modes and evaluates the method on a representative set of multimodal reasoning benchmarks. Although the formulation can extend to more modes, multi-level reasoning budgets and tool-augmented settings remain future work. Offline optimization depends on the quality and coverage of the fixed response bank and does not provide unconstrained exploration beyond its reward evidence. The slow reference supplies successful reasoning trajectories; matched-bank comparisons isolate the objective's relative gain but do not remove this dependence.

Approximate mode-format separation is used for the efficient marginal-ratio approximation, rather than the exact dual-reference objective. When formats overlap, exact correction or additional filtering may be preferable. Format-based slow-ratio diagnostics may also be ambiguous for blended outputs. The held-out transfer benchmarks may share task types or sources with ViRL39K, so these results do not establish strict OOD generalization. Finally, our efficiency comparison is measured under our reproduced setup, and absolute runtime may vary with hardware, inference engine, rollout length, and implementation details.

\section{Proof of Single-Reference BOLT}
\label{app:single_ref_bolt}

\subsection{Closed-form optimal policy}

For a fixed input \(x\), consider the KL-regularized objective
\begin{equation}
    J(\pi)
    =
    \sum_y \pi(y|x) r(x,y)
    -
    \beta
    \sum_y
    \pi(y|x)
    \log
    \frac{\pi(y|x)}{\piRef(y|x)}.
    \label{eq:app_single_ref_obj}
\end{equation}
We optimize \(J(\pi)\) subject to the normalization constraint
\(\sum_y \pi(y|x)=1\). 
The Lagrangian is
\begin{equation}
    \mathcal{J}(\pi,\lambda)
    =
    \sum_y
    \pi(y|x)
    \left[
        r(x,y)
        -
        \beta
        \log
        \frac{\pi(y|x)}{\piRef(y|x)}
    \right]
    +
    \lambda
    \left(
        \sum_y \pi(y|x)-1
    \right).
\end{equation}
Taking the derivative with respect to \(\pi(y|x)\) gives
\begin{equation}
    r(x,y)
    -
    \beta
    \left(
        \log \pi(y|x)
        -
        \log \piRef(y|x)
        +
        1
    \right)
    +
    \lambda
    =
    0.
\end{equation}
Therefore,
\begin{equation}
    \pi(y|x)
    =
    C(x)
    \piRef(y|x)
    \exp(r(x,y)/\beta),
\end{equation}
where \(C(x)\) is a normalization constant. 
Using \(\sum_y \pi(y|x)=1\), we obtain
\begin{equation}
    C(x)
    =
    \frac{1}{
        \sum_y
        \piRef(y|x)
        \exp(r(x,y)/\beta)
    }.
\end{equation}
Thus,
\begin{equation}
    \piStar(y|x)
    =
    \frac{
        \piRef(y|x)
        \exp(r(x,y)/\beta)
    }{
        Z(x)
    },
    \label{eq:app_single_ref_optimal}
\end{equation}
where
\begin{equation}
    Z(x)
    =
    \sum_y
    \piRef(y|x)
    \exp(r(x,y)/\beta)
    =
    \E_{y\sim\piRef(\cdot|x)}
    \left[
        \exp(r(x,y)/\beta)
    \right].
\end{equation}

\subsection{Forward KL to weighted SFT}

The forward-KL objective is
\begin{equation}
    \min_\theta
    \E_x
    \left[
    \KL
    \left(
        \piStar(\cdot|x)
        \|
        \pi_\theta(\cdot|x)
    \right)
    \right].
\end{equation}
Expanding the KL divergence gives
\begin{equation}
    \KL
    \left(
        \piStar(\cdot|x)
        \|
        \pi_\theta(\cdot|x)
    \right)
    =
    \E_{y\sim\piStar(\cdot|x)}
    \left[
        \log \piStar(y|x)
        -
        \log \pi_\theta(y|x)
    \right].
\end{equation}
The first term does not depend on \(\theta\). 
Thus, the objective is equivalent to minimizing
\begin{equation}
    -
    \E_x
    \E_{y\sim\piStar(\cdot|x)}
    \left[
        \log \pi_\theta(y|x)
    \right].
\end{equation}
Substituting Eq.~\eqref{eq:app_single_ref_optimal} yields
\begin{equation}
    -
    \E_x
    \E_{y\sim\piRef(\cdot|x)}
    \left[
        \frac{
            \exp(r(x,y)/\beta)
        }{
            Z(x)
        }
        \log \pi_\theta(y|x)
    \right],
\end{equation}
which is the weighted SFT objective in Eq.~\eqref{eq:bolt_single_ref}.

\section{Proofs for Dual-Reference BOLT}
\label{app:dr_bolt_derivations}

\subsection{Proof of Theorem~\ref{thm:dr_optimal}}
\label{app:proof_dr_optimal}

For a fixed input \(x\), the dual-reference KL-regularized objective is
\begin{equation}
    J(\pi)
    =
    \sum_y \pi(y|x)r(x,y)
    -
    \beta
    \sum_y
    \pi(y|x)
    \log
    \frac{\pi(y|x)}{\rhoDR(y|x)}.
    \label{eq:app_dr_obj}
\end{equation}
We optimize \(J(\pi)\) subject to \(\sum_y \pi(y|x)=1\). 
The Lagrangian is
\begin{equation}
    \mathcal{J}(\pi,\lambda)
    =
    \sum_y
    \pi(y|x)
    \left[
        r(x,y)
        -
        \beta
        \log
        \frac{\pi(y|x)}{\rhoDR(y|x)}
    \right]
    +
    \lambda
    \left(
        \sum_y \pi(y|x)-1
    \right).
\end{equation}
Taking the derivative with respect to \(\pi(y|x)\) gives
\begin{equation}
    r(x,y)
    -
    \beta
    \left(
        \log\pi(y|x)-\log\rhoDR(y|x)+1
    \right)
    +
    \lambda
    =
    0.
\end{equation}
Therefore,
\begin{equation}
    \pi(y|x)
    =
    C(x)\rhoDR(y|x)
    \exp(r(x,y)/\beta).
\end{equation}
The normalization condition gives
\begin{equation}
    C(x)
    =
    \frac{1}{
        \sum_y \rhoDR(y|x)\exp(r(x,y)/\beta)
    }.
\end{equation}
Thus,
\begin{equation}
    \piStar_{\mathrm{DR}}(y|x)
    =
    \frac{
        \rhoDR(y|x)\exp(r(x,y)/\beta)
    }{
        Z_\rho(x)
    },
\end{equation}
where
\begin{equation}
    Z_\rho(x)
    =
    \sum_y \rhoDR(y|x)\exp(r(x,y)/\beta)
    =
    \E_{y\sim\rhoDR(\cdot|x)}
    \left[
        \exp(r(x,y)/\beta)
    \right].
\end{equation}
This proves Theorem~\ref{thm:dr_optimal}.

\subsection{Proof of Proposition~\ref{prop:dr_weighted_sft}}
\label{app:proof_dr_weighted_sft}

The forward KL objective is
\begin{equation}
    \min_\theta
    \E_x
    \left[
        \KL
        \left(
            \piStar_{\mathrm{DR}}(\cdot|x)
            \|
            \pi_\theta(\cdot|x)
        \right)
    \right].
\end{equation}
Expanding the KL divergence gives
\begin{equation}
    \KL
    \left(
        \piStar_{\mathrm{DR}}(\cdot|x)
        \|
        \pi_\theta(\cdot|x)
    \right)
    =
    \E_{y\sim\piStar_{\mathrm{DR}}(\cdot|x)}
    \left[
        \log \piStar_{\mathrm{DR}}(y|x)
        -
        \log \pi_\theta(y|x)
    \right].
\end{equation}
The first term is independent of \(\theta\). 
Therefore, minimizing the forward KL is equivalent to minimizing
\begin{equation}
    -
    \E_x
    \E_{y\sim\piStar_{\mathrm{DR}}(\cdot|x)}
    \left[
        \log \pi_\theta(y|x)
    \right].
\end{equation}
Using
\begin{equation}
    \piStar_{\mathrm{DR}}(y|x)
    =
    \frac{
        \rhoDR(y|x)\exp(r(x,y)/\beta)
    }{
        Z_\rho(x)
    },
\end{equation}
we obtain
\begin{equation}
    -
    \E_x
    \E_{y\sim\rhoDR(\cdot|x)}
    \left[
        \frac{\exp(r(x,y)/\beta)}{Z_\rho(x)}
        \log \pi_\theta(y|x)
    \right],
\end{equation}
which gives Eq.~\eqref{eq:rho_loss}.

If samples are instead drawn from \(\muGen\), then for any function \(g(x,y)\),
\begin{equation}
    \E_{y\sim\rhoDR(\cdot|x)}[g(x,y)]
    =
    \E_{y\sim\muGen(\cdot|x)}
    \left[
        \frac{\rhoDR(y|x)}{\muGen(y|x)}g(x,y)
    \right],
    \label{eq:app_change_measure}
\end{equation}
assuming \(\rhoDR\) is absolutely continuous with respect to \(\muGen\). 
Taking
\begin{equation}
    g(x,y)
    =
    \frac{\exp(r(x,y)/\beta)}{Z_\rho(x)}
    \log\pi_\theta(y|x)
\end{equation}
gives Eq.~\eqref{eq:mu_loss} with
\begin{equation}
    \kappa(x,y)
    =
    \frac{\rhoDR(y|x)}{\muGen(y|x)}.
\end{equation}
This proves Proposition~\ref{prop:dr_weighted_sft}.

\subsection{Proof of Corollary~\ref{cor:routing}}
\label{app:proof_routing}

Define the joint reference distribution over mode and response as
\begin{equation}
    \rhoDR(z,y|x)
    =
    \begin{cases}
        \alpha(x)\piF(y|x), & z=\mathrm{f},\\
        \bigl(1-\alpha(x)\bigr)\piS(y|x), & z=\mathrm{s}.
    \end{cases}
\end{equation}
The reward-reweighted joint optimal policy is
\begin{equation}
    \piStar_{\mathrm{DR}}(z,y|x)
    =
    \frac{
        \rhoDR(z,y|x)\exp(r(x,y)/\beta)
    }{
        Z_\rho(x)
    }.
\end{equation}
For the fast mode,
\begin{align}
    P^\star(z=\mathrm{f}|x)
    &=
    \sum_y
    \piStar_{\mathrm{DR}}(z=\mathrm{f},y|x) \\
    &=
    \sum_y
    \frac{
        \alpha(x)\piF(y|x)\exp(r(x,y)/\beta)
    }{
        Z_\rho(x)
    } \\
    &=
    \frac{
        \alpha(x)
        \sum_y
        \piF(y|x)\exp(r(x,y)/\beta)
    }{
        Z_\rho(x)
    } \\
    &=
    \frac{
        \alpha(x)Z_{\mathrm{f}}(x)
    }{
        \alpha(x)Z_{\mathrm{f}}(x)
        +
        \bigl(1-\alpha(x)\bigr)Z_{\mathrm{s}}(x)
    }.
\end{align}
The slow-mode probability follows by normalization:
\begin{equation}
    P^\star(z=\mathrm{s}|x)
    =
    1-P^\star(z=\mathrm{f}|x).
\end{equation}
This proves Corollary~\ref{cor:routing}.

\subsection{Proof of Proposition~\ref{prop:modewise}}
\label{app:proof_modewise}

The exact behavior-reference correction is
\begin{equation}
    \kappa(x,y)
    =
    \frac{
        \rhoDR(y|x)
    }{
        \muGen(y|x)
    }
    =
    \frac{
        \alpha(x)\piF(y|x)
        +
        \bigl(1-\alpha(x)\bigr)\piS(y|x)
    }{
        q(x)\piF(y|x)
        +
        \bigl(1-q(x)\bigr)\piS(y|x)
    }.
    \label{eq:app_exact_kappa}
\end{equation}

\paragraph{Fast samples.}
For a fast sample \(y_{\mathrm{f}}\), define
\begin{equation}
    \delta_{\mathrm{f}}(x,y_{\mathrm{f}})
    =
    \frac{\piS(y_{\mathrm{f}}|x)}{\piF(y_{\mathrm{f}}|x)}.
\end{equation}
Dividing the numerator and denominator of Eq.~\eqref{eq:app_exact_kappa} by
\(\piF(y_{\mathrm{f}}|x)\), we obtain
\begin{equation}
    \kappa(x,y_{\mathrm{f}})
    =
    \frac{
        \alpha(x)+\bigl(1-\alpha(x)\bigr)\delta_{\mathrm{f}}
    }{
        q(x)+\bigl(1-q(x)\bigr)\delta_{\mathrm{f}}
    }.
    \label{eq:app_kappa_fast}
\end{equation}
Subtracting the mode-wise constant \(\alpha(x)/q(x)\) gives
\begin{align}
    \kappa(x,y_{\mathrm{f}})
    -
    \frac{\alpha(x)}{q(x)}
    &=
    \frac{
        \alpha(x)+\bigl(1-\alpha(x)\bigr)\delta_{\mathrm{f}}
    }{
        q(x)+\bigl(1-q(x)\bigr)\delta_{\mathrm{f}}
    }
    -
    \frac{\alpha(x)}{q(x)} \\
    &=
    \frac{
        \delta_{\mathrm{f}}
        \left[
            q(x)-\alpha(x)
        \right]
    }{
        q(x)
        \left[
            q(x)+\bigl(1-q(x)\bigr)\delta_{\mathrm{f}}
        \right]
    }.
    \label{eq:app_fast_residual_exact}
\end{align}
By Assumption~\ref{assump:sep}, \(\delta_{\mathrm{f}}\leq\varepsilon_{\mathrm{f}}\). 
If \(q(x)\in[\eta,1-\eta]\), the denominator in Eq.~\eqref{eq:app_fast_residual_exact} is bounded away from zero. 
Therefore,
\begin{equation}
    \kappa(x,y_{\mathrm{f}})
    =
    \frac{\alpha(x)}{q(x)}
    +
    O(\varepsilon_{\mathrm{f}}).
\end{equation}

\paragraph{Slow samples.}
For a slow sample \(y_{\mathrm{s}}\), define
\begin{equation}
    \delta_{\mathrm{s}}(x,y_{\mathrm{s}})
    =
    \frac{\piF(y_{\mathrm{s}}|x)}{\piS(y_{\mathrm{s}}|x)}.
\end{equation}
Dividing the numerator and denominator of Eq.~\eqref{eq:app_exact_kappa} by
\(\piS(y_{\mathrm{s}}|x)\), we obtain
\begin{equation}
    \kappa(x,y_{\mathrm{s}})
    =
    \frac{
        1-\alpha(x)+\alpha(x)\delta_{\mathrm{s}}
    }{
        1-q(x)+q(x)\delta_{\mathrm{s}}
    }.
    \label{eq:app_kappa_slow}
\end{equation}
Subtracting the mode-wise constant \((1-\alpha(x))/(1-q(x))\) gives
\begin{align}
    \kappa(x,y_{\mathrm{s}})
    -
    \frac{1-\alpha(x)}{1-q(x)}
    &=
    \frac{
        1-\alpha(x)+\alpha(x)\delta_{\mathrm{s}}
    }{
        1-q(x)+q(x)\delta_{\mathrm{s}}
    }
    -
    \frac{1-\alpha(x)}{1-q(x)} \\
    &=
    \frac{
        \delta_{\mathrm{s}}
        \left[
            \alpha(x)-q(x)
        \right]
    }{
        \bigl(1-q(x)\bigr)
        \left[
            1-q(x)+q(x)\delta_{\mathrm{s}}
        \right]
    }.
    \label{eq:app_slow_residual_exact}
\end{align}
By Assumption~\ref{assump:sep}, \(\delta_{\mathrm{s}}\leq\varepsilon_{\mathrm{s}}\). 
If \(q(x)\in[\eta,1-\eta]\), the denominator in Eq.~\eqref{eq:app_slow_residual_exact} is bounded away from zero. 
Therefore,
\begin{equation}
    \kappa(x,y_{\mathrm{s}})
    =
    \frac{1-\alpha(x)}{1-q(x)}
    +
    O(\varepsilon_{\mathrm{s}}).
\end{equation}
This proves Proposition~\ref{prop:modewise}.
\section{Empirical Validation of Dual-Reference Assumptions}
\label{app:dual_ref_validation}

\subsection{Validation of Mode-Format Separation}
\label{app:support_validation}

The pointwise approximation to the marginal correction in Proposition~\ref{prop:modewise} relies on the smallness of the cross-mode ratios
\[
    \delta_{\mathrm{f}}(x,y_{\mathrm{f}})
    =
    \frac{\piS(y_{\mathrm{f}}|x)}{\piF(y_{\mathrm{f}}|x)},
    \quad
    \delta_{\mathrm{s}}(x,y_{\mathrm{s}})
    =
    \frac{\piF(y_{\mathrm{s}}|x)}{\piS(y_{\mathrm{s}}|x)}.
\]
We validate this assumption with teacher-forced cross-likelihood evaluation. 
For each side, we sample 1000 completions and compute their log-likelihoods under both the fast policy \(\piF\) and the slow policy \(\piS\). 
We run the evaluation on the main generation path and on an ablation path. 
The key diagnostic is whether the off-diagonal likelihoods are exponentially smaller than the diagonal likelihoods.

\begin{figure}[h]
\centering
\includegraphics[width=\linewidth]{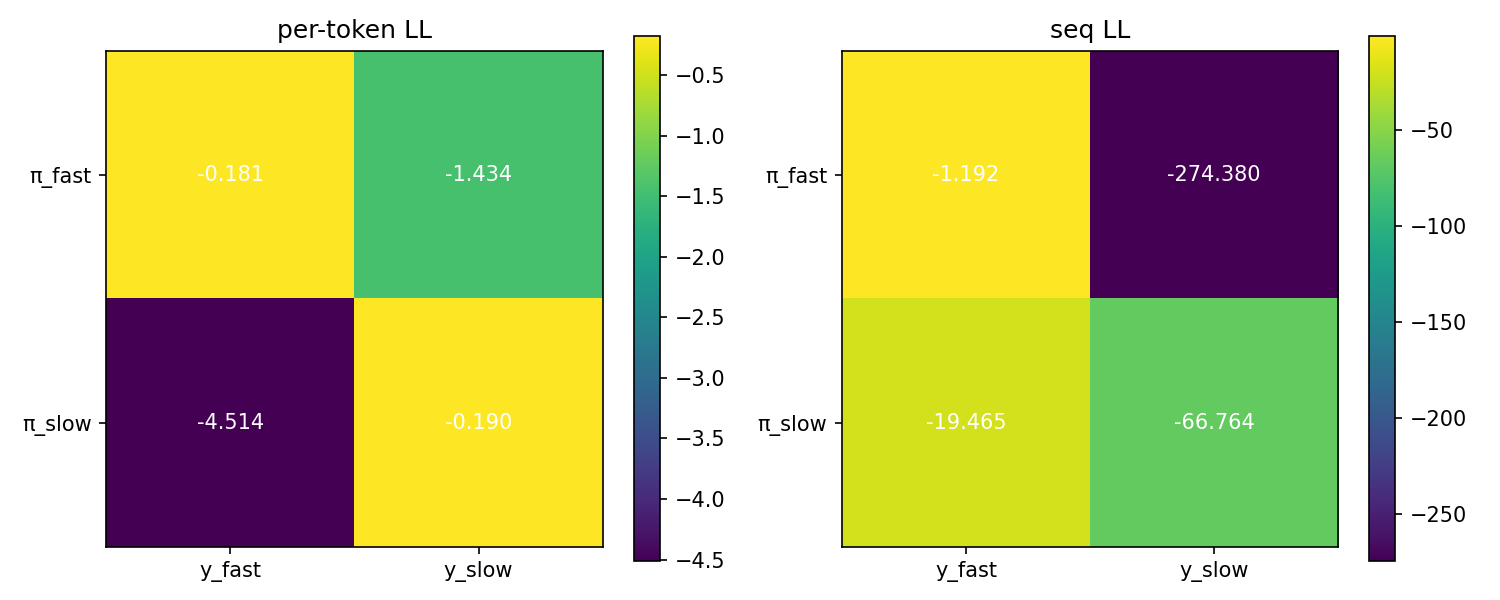}
\caption{
Cross-likelihood heatmaps on the main path. 
The left panel reports mean per-token log-likelihood; the right panel reports mean sequence log-likelihood. 
Diagonal entries correspond to in-mode likelihoods, while off-diagonal entries measure cross-mode likelihoods.
}
\label{fig:cross_likelihood_main}
\end{figure}

\paragraph{Main-path result.}
Figure~\ref{fig:cross_likelihood_main} shows the main-path cross-likelihood heatmaps, and Table~\ref{tab:cross_likelihood_main} reports the numerical values. 
Each cell reports mean per-token log-likelihood followed by mean sequence log-likelihood in nats.

\begin{table}[h]
\centering
\small
\caption{
Cross-likelihood matrix on the main path. Each entry reports mean per-token log-likelihood / mean sequence log-likelihood in nats.
}
\label{tab:cross_likelihood_main}
\begin{tabular}{lcc}
\toprule
 & \(y_{\mathrm{f}}\) \((\bar L=6.21)\) & \(y_{\mathrm{s}}\) \((\bar L=276.64)\) \\
\midrule
\(\piF\) & \(\mathbf{-0.181 / -1.192}\) & \(-1.434 / \mathbf{-274.380}\) \\
\(\piS\) & \(-4.514 / \mathbf{-19.465}\) & \(\mathbf{-0.190 / -66.764}\) \\
\bottomrule
\end{tabular}
\end{table}

The diagonal per-token log-likelihoods are close to zero, around \(-0.19\), indicating that each policy assigns high confidence to its own response family. 
The off-diagonal sequence likelihoods are much smaller. 
For fast samples, the relative sequence log-likelihood ratio is
\begin{equation}
    \log \frac{\piS(y_{\mathrm{f}}|x)}{\piF(y_{\mathrm{f}}|x)}
    =
    -19.465 - (-1.192)
    =
    -18.273.
\end{equation}
For slow samples, the relative sequence log-likelihood ratio is
\begin{equation}
    \log \frac{\piF(y_{\mathrm{s}}|x)}{\piS(y_{\mathrm{s}}|x)}
    =
    -274.380 - (-66.764)
    =
    -207.616.
\end{equation}
Thus, the cross-mode ratios are on the order of \(e^{-18}\) for \(\piS(y_{\mathrm{f}}|x)/\piF(y_{\mathrm{f}}|x)\) and \(e^{-208}\) for \(\piF(y_{\mathrm{s}}|x)/\piS(y_{\mathrm{s}}|x)\). 
These values strongly support the approximate mode-format separation assumption used in Proposition~\ref{prop:modewise}.

\paragraph{Two mechanisms behind approximate separation.}
The two off-diagonal directions are separated for different reasons. 
Table~\ref{tab:separation_mechanisms} summarizes the mechanisms.

\begin{table}[h]
\centering
\small
\caption{
Different mechanisms behind the two cross-mode likelihood gaps.
}
\label{tab:separation_mechanisms}
\begin{tabular}{@{}p{0.22\linewidth}p{0.52\linewidth}p{0.20\linewidth}@{}}
\toprule
Direction & Mechanism & Length dependence \\
\midrule
\(\piF(y_{\mathrm{s}})\) 
& \textbf{Length-compounded decay.} The fast policy assigns slightly lower probability to many reasoning tokens; over long traces this compounds into an exponentially small sequence ratio. Empirically, the per-token mismatch is amplified over \(\bar L\approx277\) tokens. 
& Strong: \(r(L,\Delta_{\mathrm{seq}})=-0.902\). \\
\(\piS(y_{\mathrm{f}})\) 
& \textbf{Prefix bottleneck.} The slow policy tends to start with a reasoning-format prefix such as \texttt{<think>}; direct fast answers start with answer tokens instead. The mismatch is therefore concentrated near the first token rather than accumulated over length. 
& Weak: \(r(L,\Delta_{\mathrm{seq}})=+0.226\). \\
\bottomrule
\end{tabular}
\end{table}

This asymmetry is important. 
Approximate separation is not merely a length effect. 
For \(\piF(y_{\mathrm{s}}|x)\), length amplifies the mismatch over long reasoning traces. 
For \(\piS(y_{\mathrm{f}}|x)\), the dominant cause is a prefix-level format bottleneck: after the assistant header, the slow policy strongly prefers a reasoning marker, while \(y_{\mathrm{f}}\) begins directly with the answer. 
This supports our use of the weaker \(\varepsilon_{\mathrm{f}},\varepsilon_{\mathrm{s}}\)-separation assumption instead of a symmetric length-dependent assumption.

\begin{figure}[h]
\centering
\includegraphics[width=\linewidth]{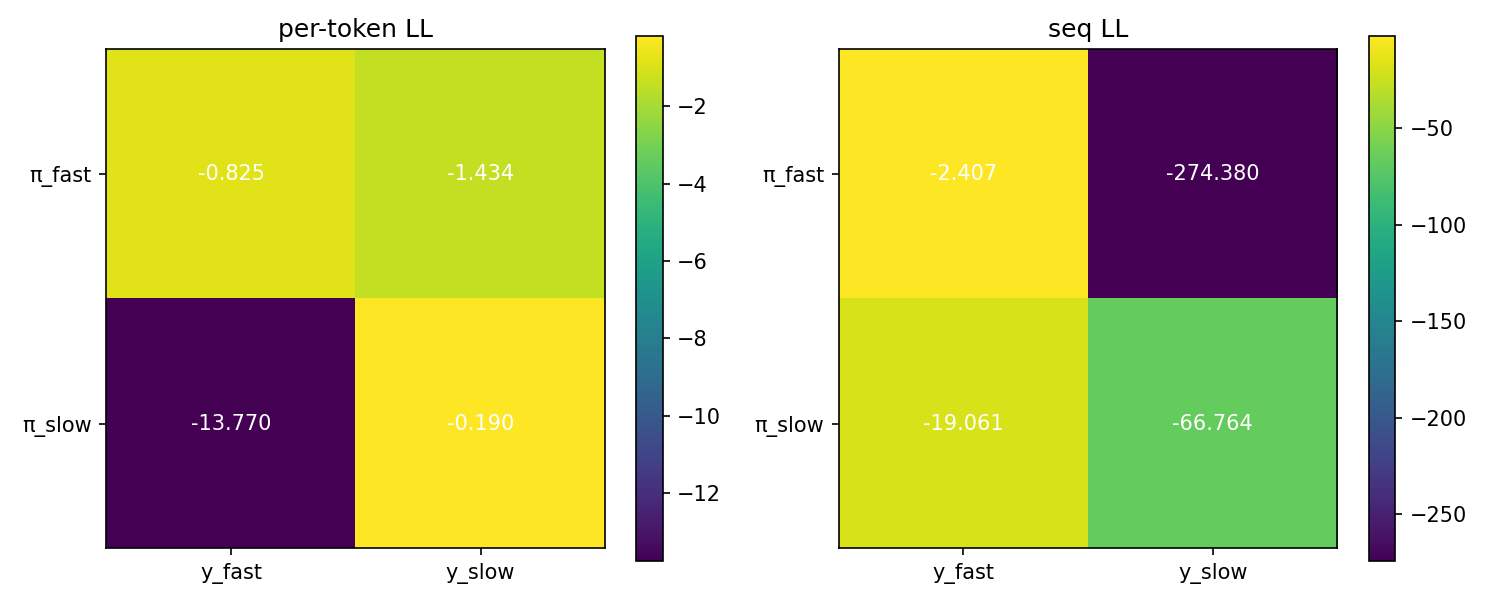}
\caption{
Cross-likelihood heatmaps on the ablation path. 
The slow policy still assigns low likelihood to direct fast answers, indicating that the prefix bottleneck is a structural property of the checkpoint. 
A small fraction of contaminated fast samples weaken the fast-pool purity.
}
\label{fig:cross_likelihood_ablation}
\end{figure}

\paragraph{Ablation path.}
We further evaluate an ablation path to check whether the slow-to-fast gap is merely a data-source artifact. 
Figure~\ref{fig:cross_likelihood_ablation} shows the ablation-path heatmaps.

The sequence log-likelihood of \(\piS(y_{\mathrm{f}}|x)\) changes only slightly, from \(-19.465\) on the main path to approximately \(-19.07\) on the ablation path. 
This shows that the prefix bottleneck is a checkpoint-level structural feature and cannot be removed simply by changing the source of \(y_{\mathrm{f}}\). 
However, the ablation fast pool contains a small number of reasoning-style long answers: about \(3\%\)--\(5\%\) of samples bypass the disable-thinking prefix and generate reasoning-style completions, with the longest reaching \(L=216\). 
These contaminated samples weaken the strict separation of the fast pool and motivate length- or marker-based filtering, such as removing fast samples with reasoning markers or length exceeding a threshold such as \(L>30\) generated tokens. Exact marginal correction is another fallback. These are suggested controls for contaminated candidates, rather than a demonstrated general robustness guarantee.

\paragraph{Conclusion.}
The cross-likelihood results support three conclusions. 
First, the approximate mode-format separation assumption holds strongly on the main path: diagonal probabilities are high, while off-diagonal sequence likelihood ratios fall in the \(e^{-18}\) to \(e^{-208}\) range. 
Second, the separation mechanisms are asymmetric and structural: fast-to-slow mismatch is length-compounded, while slow-to-fast mismatch is caused by a prefix bottleneck. 
Third, ablation-path failures are driven by contamination in the fast pool rather than by a failure of the dual-reference assumption itself.

\subsection{Validation of Mode-wise Correction}
\label{app:kappa_validation}

We next validate whether the exact behavior-reference correction
\[
    \kappa_{\mathrm{exact}}(x,y)
    =
    \frac{\rhoDR(y|x)}{\muGen(y|x)}
\]
can be replaced by the mode-wise correction derived in Proposition~\ref{prop:modewise}. 
We reuse the log-likelihood files from Appendix~\ref{app:support_validation}, join samples by their sample IDs, and obtain \(n=1958\) samples on the main path and \(n=1964\) samples on the ablation path. 
For each sample, we compare three quantities:
\begin{align}
    \log w_{\mathrm{exact}}(x,y)
    &=
    \log
    \frac{
        \alpha\piF(y|x)+(1-\alpha)\piS(y|x)
    }{
        q\piF(y|x)+(1-q)\piS(y|x)
    }, \\
    \log w_{\mathrm{theory}}(x,y_{\mathrm{f}})
    &=
    \log \frac{\alpha}{q}, \\
    \log w_{\mathrm{theory}}(x,y_{\mathrm{s}})
    &=
    \log \frac{1-\alpha}{1-q}.
\end{align}
As a diagnostic of single-reference failure, we also compute
\begin{equation}
    \log w_{\mathrm{single}}(x,y)
    =
    \log \piF(y|x)-\log \piS(y|x).
\end{equation}
This quantity shows how severely a single reference can suppress samples from the other mode.

\begin{figure}[h]
\centering
\includegraphics[width=\linewidth]{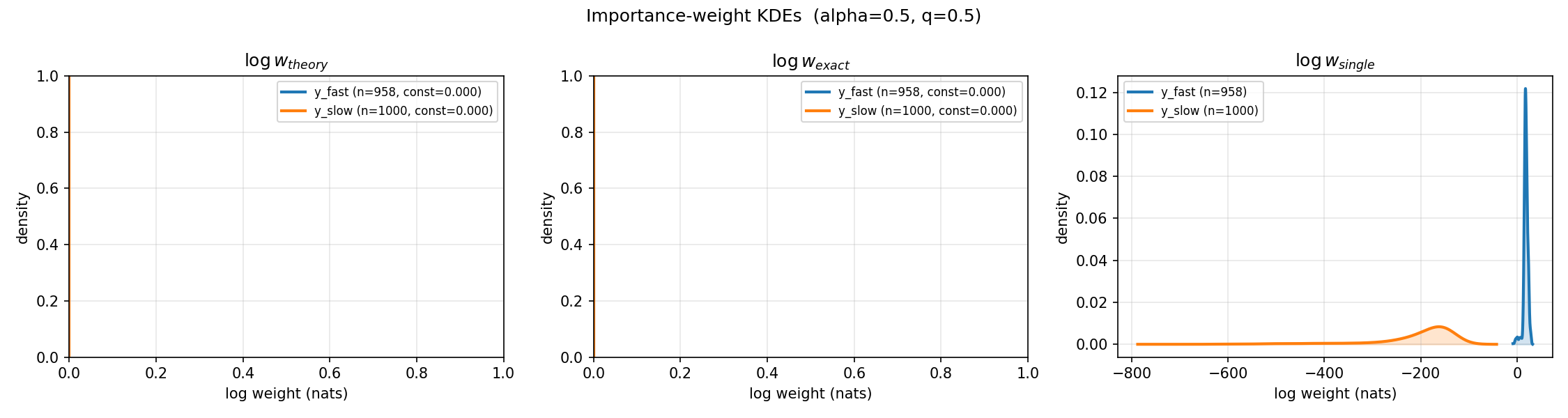}
\caption{
Correction weights in the balanced setting \(\alpha=q=0.5\). 
Both \(\log w_{\mathrm{theory}}\) and \(\log w_{\mathrm{exact}}\) collapse to zero, so correlation is undefined. 
The single-reference diagnostic \(\log w_{\mathrm{single}}\) still shows a strong bimodal structure.
}
\label{fig:kappa_degenerate}
\end{figure}

\paragraph{Balanced case: \(\alpha=q=0.5\).}
When \(\alpha=q=0.5\), the desired reference mixture and the empirical sampling mixture are identical:
\[
    \rhoDR=\muGen.
\]
Therefore,
\[
    \kappa_{\mathrm{exact}}(x,y)=1,
    \quad
    \log w_{\mathrm{exact}}(x,y)=0.
\]
The mode-wise correction also gives
\[
    \log w_{\mathrm{theory}}=0
\]
for both fast and slow samples. 
Thus, all samples satisfy
\[
    \log w_{\mathrm{theory}}=\log w_{\mathrm{exact}}=0.
\]
This is a useful sanity check, but it cannot validate correlation because both variables have zero variance. 
Figure~\ref{fig:kappa_degenerate} shows this degenerate case.

Although \(w_{\mathrm{theory}}\) and \(w_{\mathrm{exact}}\) are degenerate in this setting, \(w_{\mathrm{single}}\) still reveals the failure mode of single-reference weighting. 
Table~\ref{tab:single_weight_stats} reports the statistics.

\begin{table}[h]
\centering
\small
\caption{
Statistics of \(\log w_{\mathrm{single}}=\log\piF(y|x)-\log\piS(y|x)\) in the balanced setting.
}
\label{tab:single_weight_stats}
\begin{tabular}{lcccc}
\toprule
Samples & Mean & Std & Min & Max \\
\midrule
\(\log w_{\mathrm{single}}\mid y_{\mathrm{f}}\) \((n=958)\) 
& \(+18.27\) & \(4.76\) & \(-7.96\) & \(+29.49\) \\
\(\log w_{\mathrm{single}}\mid y_{\mathrm{s}}\) \((n=1000)\) 
& \(-207.62\) & \(91.96\) & \(-769.90\) & \(-101.92\) \\
\bottomrule
\end{tabular}
\end{table}

The two modes are almost disjoint on the log scale, with a gap of roughly \(100\) nats between the fast and slow regions. 
For slow samples, \(100\%\) of ratios are below \(e^{-100}\), and about \(34\%\) are below \(e^{-200}\). 
Feeding such single-reference weights directly into training would underflow in common low-precision formats and would effectively remove the slow samples from the gradient. 
This directly explains the always-fast failure mode of single-reference BOLT with a fast reference.

\begin{figure}[h]
\centering
\includegraphics[width=\linewidth]{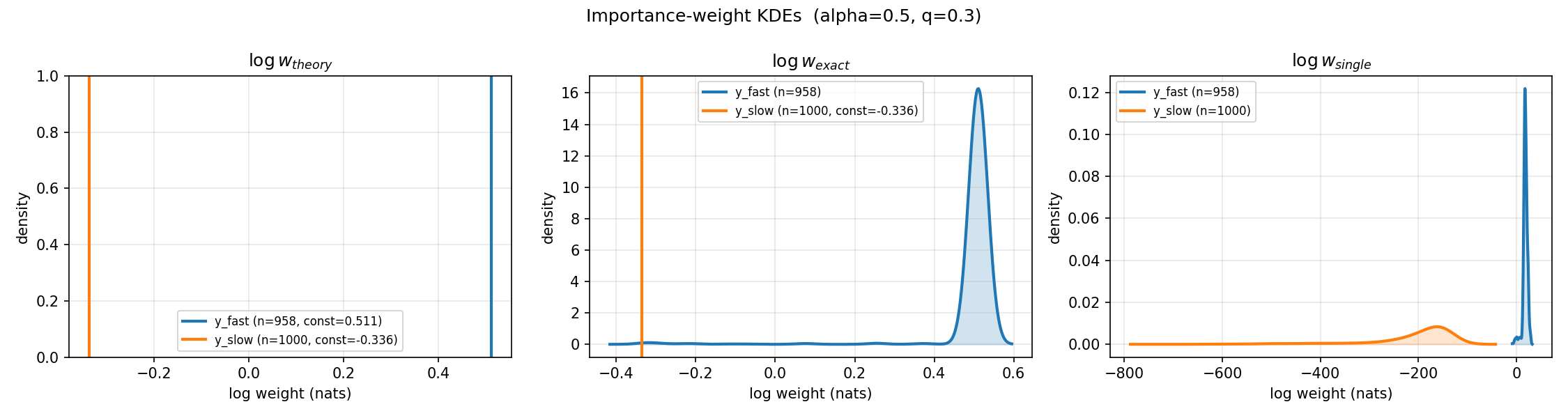}
\caption{
Correction weights on the main path with \(\alpha=0.5,q=0.3\). 
Since \(\alpha\ne q\), \(\log w_{\mathrm{theory}}\) is non-constant and can be compared with \(\log w_{\mathrm{exact}}\).
}
\label{fig:kappa_q03}
\end{figure}

\begin{figure}[h]
\centering
\includegraphics[width=\linewidth]{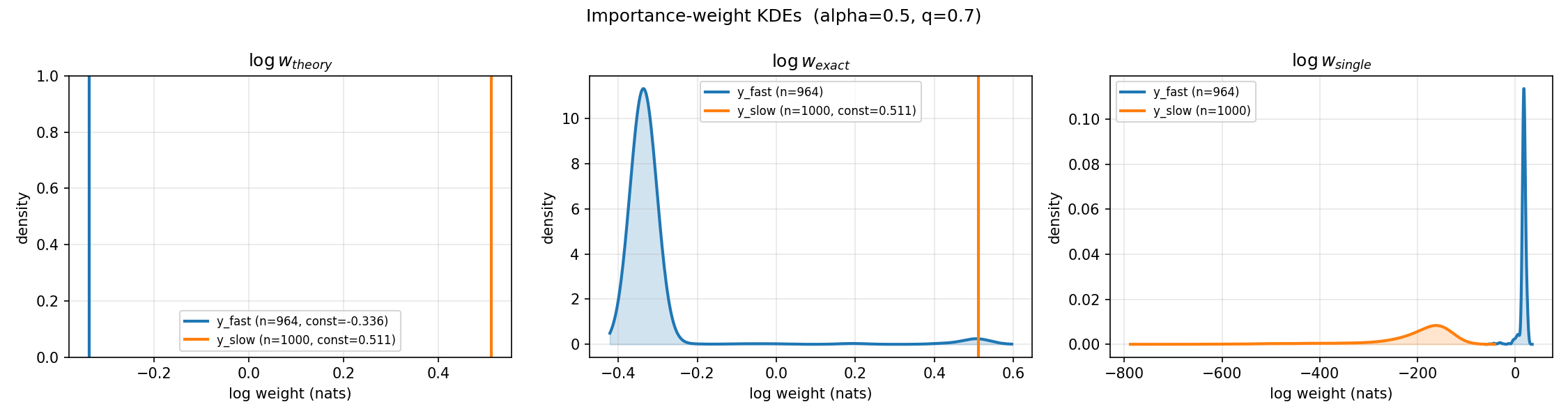}
\caption{
Correction weights on the main path with \(\alpha=0.5,q=0.7\). 
The agreement between \(w_{\mathrm{theory}}\) and \(w_{\mathrm{exact}}\) is strongest in this setting.
}
\label{fig:kappa_q07}
\end{figure}

\begin{figure}[h]
\centering
\includegraphics[width=\linewidth]{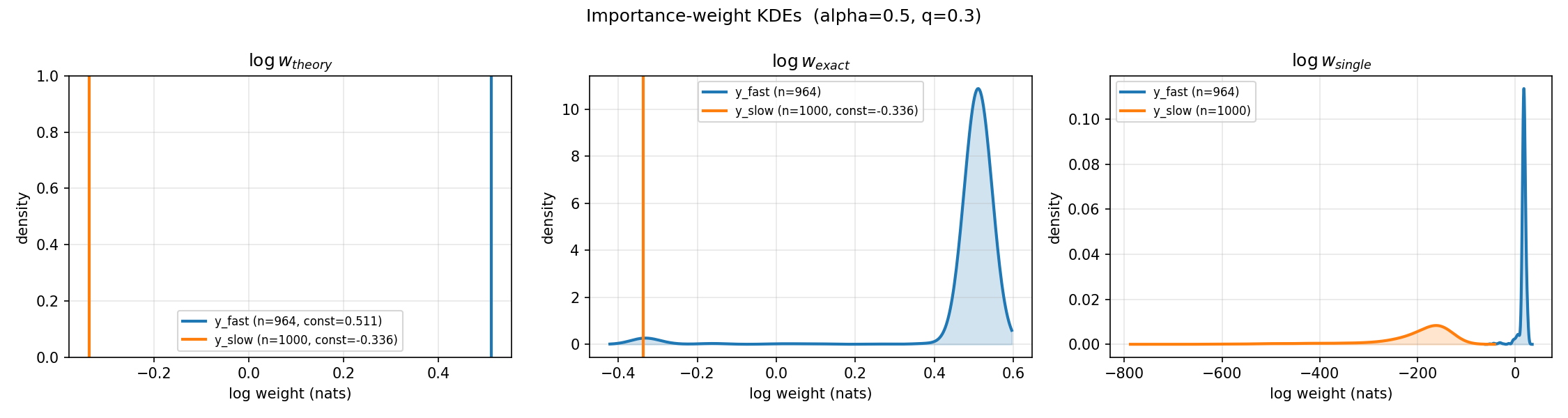}
\caption{
Correction weights on the ablation path with \(\alpha=0.5,q=0.3\). 
The equivalence degrades because the ablation fast pool contains reasoning-style contaminated samples.
}
\label{fig:kappa_ablation_q03}
\end{figure}

\paragraph{Non-degenerate settings: \(\alpha\ne q\).}
To test equivalence between \(w_{\mathrm{theory}}\) and \(w_{\mathrm{exact}}\), we need \(\alpha\ne q\). 
We evaluate two main-path settings, \((\alpha,q)=(0.5,0.3)\) and \((0.5,0.7)\), and one ablation-path setting, \((\alpha,q)=(0.5,0.3)\). 
Figures~\ref{fig:kappa_q03}, \ref{fig:kappa_q07}, and \ref{fig:kappa_ablation_q03} show the corresponding KDE and scatter plots.

The mode gap between the fast and slow constants is \(0.847\) nats. 
Table~\ref{tab:kappa_equivalence} reports the quantitative comparison.

\begin{table}[h]
\centering
\small
\caption{
Equivalence between exact correction and mode-wise correction. 
The main path shows strong agreement, while the ablation path degrades due to contaminated fast samples.
}
\label{tab:kappa_equivalence}
\begin{tabular}{lccc}
\toprule
Metric & main, \(q=0.3\) & main, \(q=0.7\) & ablation, \(q=0.3\) \\
\midrule
Pooled \(r(w_{\mathrm{theory}},w_{\mathrm{exact}})\) 
& \(0.9877\) & \(\mathbf{0.9902}\) & \(0.9730\) \\
\(|\mathrm{err}|>0.01\) 
& \(1.5\%\) & \(1.2\%\) & \(2.2\%\) \\
Max \(|\mathrm{err}|\) 
& \(0.847\) & \(0.847\) & \(0.847\) \\
Fast margin std 
& \(4.76\) & \(4.76\) & \(\mathbf{7.87}\) \\
Fast margin min 
& \(-7.96\) & \(-7.96\) & \(\mathbf{-66.38}\) \\
\bottomrule
\end{tabular}
\end{table}

The main-path results show that the mode-wise correction is quantitatively close to the exact correction. 
The \(q=0.7\) setting reaches the target \(r>0.99\), while \(q=0.3\) is slightly below the threshold but still highly correlated. 
The ablation setting degrades more noticeably, driven by the heavy left tail of the fast margin caused by reasoning-style contaminated fast samples.

\paragraph{Key quantitative structure.}
The equivalence has three useful quantitative properties. 
First, the slow side is extremely stable: the median margin is about \(177\) nats, and the ratio \(\piF(y_{\mathrm{s}}|x)/\piS(y_{\mathrm{s}}|x)\) underflows to zero even in high-precision arithmetic. 
As a result, \(w_{\mathrm{exact}}\) and \(w_{\mathrm{theory}}\) agree up to numerical precision on slow samples, with standard deviation around \(7.85\times10^{-15}\). 
Second, the fast side is controlled by the margin: the error decays exponentially with the margin, and increasing the margin by \(5\) nats reduces the error by roughly \(10^3\). 
Third, the maximum possible deviation is bounded by the mode gap of \(0.847\) nats, because \(w_{\mathrm{exact}}\) is a monotone function of the likelihood ratio and interpolates between the two mode constants \(\alpha/q\) and \((1-\alpha)/(1-q)\).

\paragraph{Conclusion.}
These results support three conclusions. 
First, the mode-wise constant approximation is quantitatively valid on the main path, although the exact \(r>0.99\) criterion is sensitive to \((\alpha,q)\). 
Second, the single-reference diagnostic \(w_{\mathrm{single}}\) exhibits an extreme dynamic range: slow samples are suppressed by more than \(200\) nats on average, which explains the numerical and optimization failure of single-reference BOLT. 
Third, equivalence degrades on the ablation path because the fast pool contains reasoning-style contaminated samples. 
This motivates cleaning the fast pool with reasoning-marker filters and length thresholds before using it as a strict dual-reference comparison.

\subsection{Optimization Dynamics of Exact, Mode-wise, and Single-Reference Weights}
\label{app:optimization_dynamics}

The previous section validates the static equivalence between the exact correction and the mode-wise correction. 
We further test whether this equivalence persists during optimization. 
We compare four training buckets under \((\alpha,q)=(0.5,0.7)\): 
\textbf{mode-wise}, which uses the theoretical mode-wise correction in Proposition~\ref{prop:modewise}; 
\textbf{exact}, which uses the exact correction \(\kappa(x,y)=\rho(y|x)/\mu(y|x)\); 
\textbf{single}, which uses a degenerate single-reference weight; 
and \textbf{mode-wise-full}, which keeps the source-wise constants without group normalization and retains all all-wrong groups. 
All buckets share the same response bank, batch order, optimizer, and LoRA setup; the only difference is the sample-weight column. 
We train for 500 steps with AdamW and evaluate implicit KL by replaying 200 held-out samples every 25 steps using teacher forcing.

\begin{figure}[h]
\centering
\includegraphics[width=\linewidth]{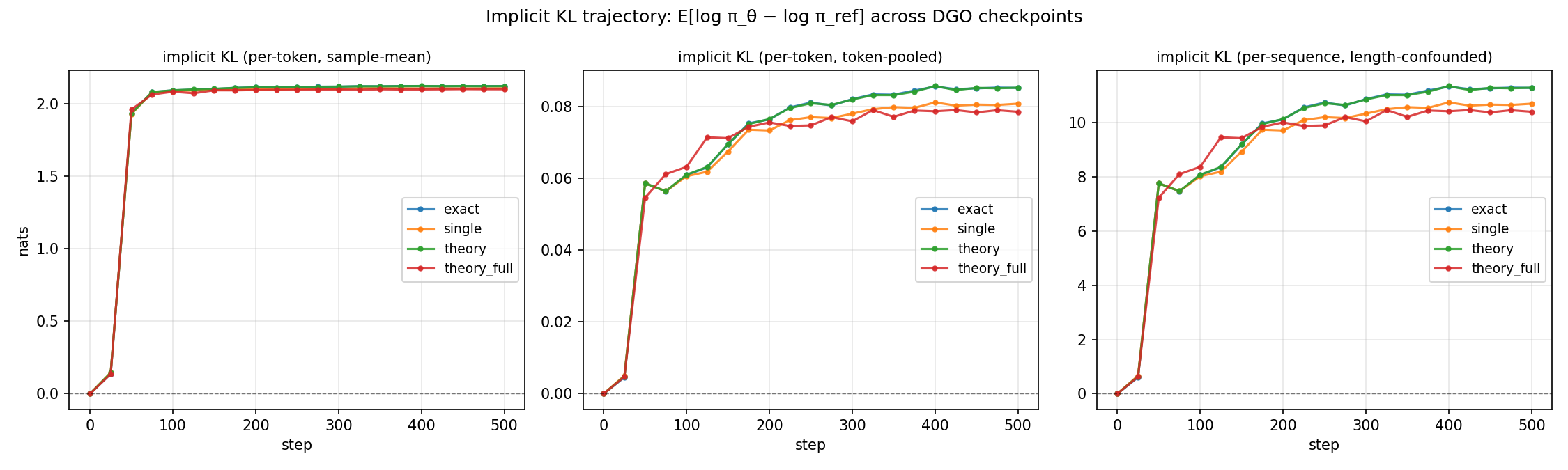}
\caption{
Implicit KL trajectory on the main path under \((\alpha,q)=(0.5,0.7)\). 
The mode-wise and exact buckets remain nearly indistinguishable throughout training.
}
\label{fig:opt_main_kl}
\end{figure}

\begin{figure}[h]
\centering
\includegraphics[width=0.48\linewidth]{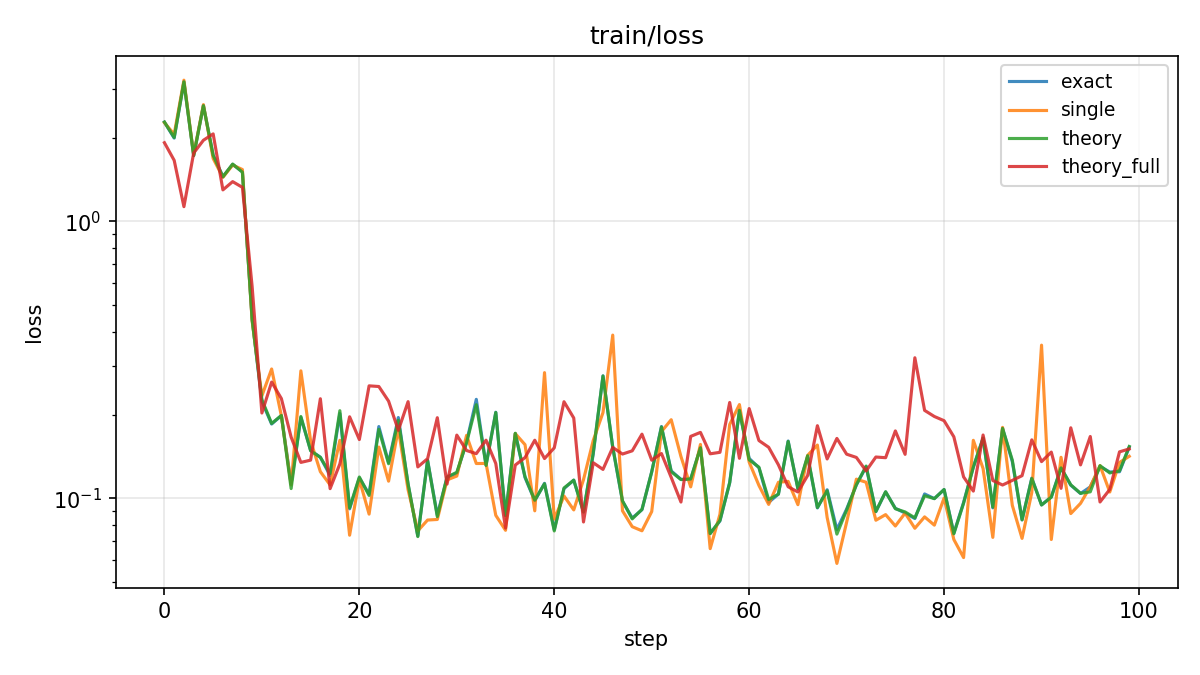}
\hfill
\includegraphics[width=0.48\linewidth]{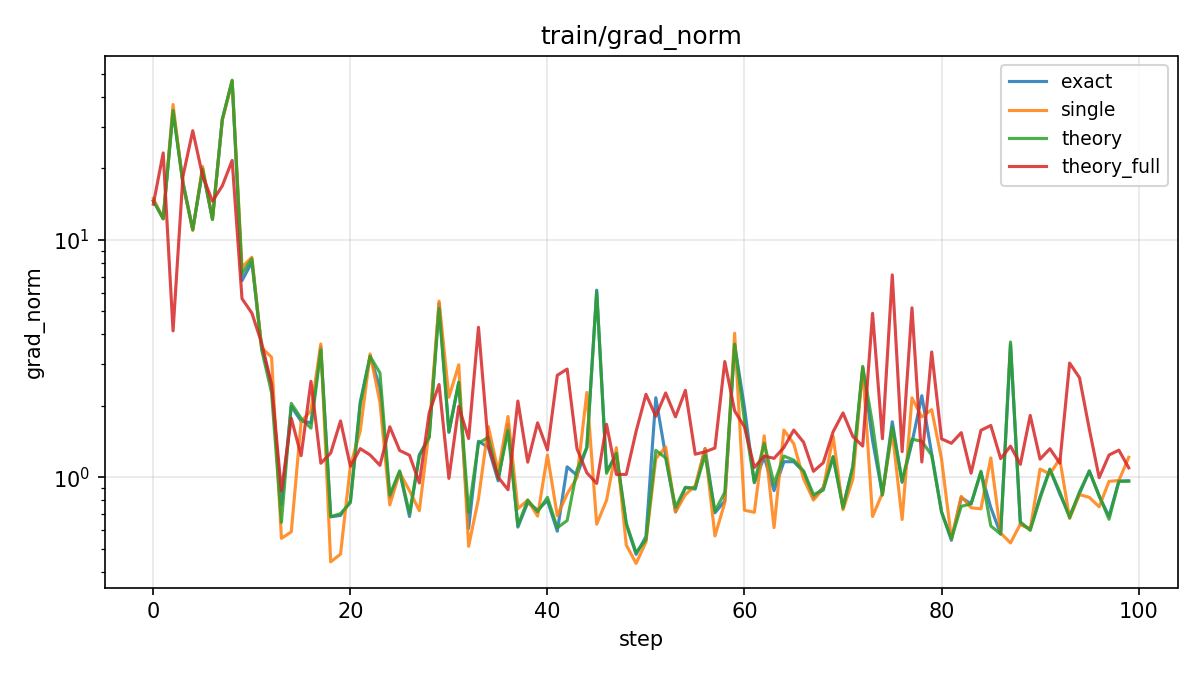}
\caption{
Training loss and gradient norm on the main path. 
The mode-wise and exact corrections have nearly identical optimization dynamics, while the full mode-wise implementation yields a smoother gradient profile.
}
\label{fig:opt_main_loss_grad}
\end{figure}

\paragraph{Main-path dynamics.}
Figure~\ref{fig:opt_main_kl} shows the implicit KL trajectory on the main path. 
Figures~\ref{fig:opt_main_loss_grad} shows the training loss and gradient norm. 
Table~\ref{tab:opt_main_dynamics} summarizes the key statistics.

\begin{table}[h]
\centering
\small
\caption{
Optimization dynamics on the main path. KL values are sequence-level implicit KL in nats.
}
\label{tab:opt_main_dynamics}
\begin{tabular}{lcccccc}
\toprule
Bucket & Step 25 & Step 50 & Step 100 & Step 500 & Final Loss & Grad Peak \\
\midrule
Mode-wise & \(0.64\) & \(7.75\) & \(8.08\) & \(\mathbf{11.28}\) & \(0.154\) & \(47.0\) \\
Exact & \(0.61\) & \(7.77\) & \(8.06\) & \(\mathbf{11.27}\) & \(0.153\) & \(46.5\) \\
Single-ref & \(0.67\) & \(7.76\) & \(8.02\) & \(10.69\) & \(0.142\) & \(46.7\) \\
Mode-wise-full & \(0.62\) & \(7.23\) & \(8.36\) & \(10.39\) & \(0.151\) & \(\mathbf{28.8}\) \\
\bottomrule
\end{tabular}
\end{table}

The mode-wise and exact buckets are almost identical: their step-wise KL difference stays below \(0.1\) nats, their gradient-norm AUC differs by less than \(1\%\), and their final losses differ by only \(0.001\). 
This shows that the static equivalence observed in Appendix~\ref{app:kappa_validation} does not accumulate into a meaningful optimization gap over 500 steps. 
In contrast, the single-reference bucket obtains a slightly lower final loss and lower final KL, but this is a misleading signal: the single-reference weight underflows for slow samples and effectively drops part of the slow-response family from the gradient. 
In our run, \(28/930\) slow rows underflow to zero in bf16, and the remaining slow rows are also assigned extremely small weights. 
Thus, the single-reference dynamics are not better optimization, but rather a collapse toward a fast-path SFT-like objective.

The mode-wise-full bucket gives the smoothest trajectory, with the gradient-norm peak reduced from roughly \(47\) to \(28.8\). 
This implementation is closer to the mathematical source-wise correction, but it uses a larger effective dataset and covers fewer epochs under the same 500-step budget. 
We therefore treat it as an informative stability check rather than the main apples-to-apples comparison.

\begin{figure}[h]
\centering
\includegraphics[width=\linewidth]{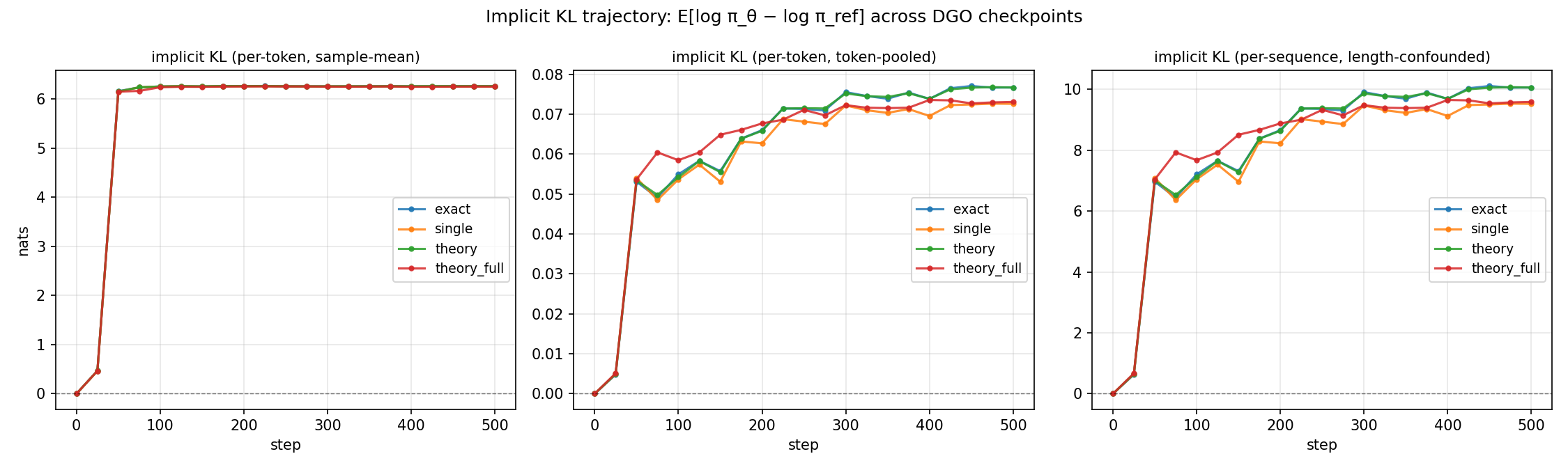}
\caption{
Implicit KL trajectory on the ablation path. 
The relative differences between buckets remain similar to the main path, but the absolute optimization scale is much larger due to first-token mismatch.
}
\label{fig:opt_ablation_kl}
\end{figure}

\paragraph{Ablation-path dynamics.}
We repeat the same comparison on the ablation path. 
Figure~\ref{fig:opt_ablation_kl} shows the implicit KL trajectory.

The bucket-relative behavior is similar to the main path, but the absolute scale is much larger. 
For the mode-wise bucket, KL per token at step 500 increases from \(2.121\) on the main path to \(6.261\) on the ablation path; the initial training loss increases from \(2.28\) to \(6.45\); and the peak gradient norm increases from \(47.0\) to \(128.4\). 
This is caused by a stronger base-data mismatch: the base model is a thinking checkpoint, while ablation-path fast samples are extremely short and require the model to move the first output token away from the reasoning prefix. 
This directly exposes the first-token bottleneck during optimization.

\paragraph{Takeaways.}
These dynamics support three conclusions. 
First, mode-wise and exact correction show closely matched static weights and optimization dynamics in this diagnostic; this is empirical agreement rather than exact finite-sample equivalence. 
Second, the single-reference objective fails by numerical underflow and gradient removal for one response family, explaining the mode-collapse behavior observed in the main experiments. 
Third, the absolute optimization difficulty is driven by base-data alignment, especially the first-token reasoning-prefix bottleneck, while the relative bucket differences are driven by the choice of sample weights.

\section{Experimental Details}
\label{app:exp_details}

\subsection{Datasets and Evaluation Protocol}
\label{app:dataset_metrics}

We evaluate \textsc{DR-BOLT} in one in-distribution (IID) setting and three held-out cross-benchmark transfer settings. 
For IID evaluation, we train on the ScienceQA training split and evaluate on the official ScienceQA test split~\citep{lu2022scienceqa}. 
For cross-benchmark transfer evaluation, we train on ViRL39K~\citep{wang2025vlrethinker} and evaluate on MathVision~\citep{wang2024mathvision}, DynaMath~\citep{zou2025dynamath}, and MMMU-Pro~\citep{yue2025mmmupro}. 
All reported scores use the official metric when available and are normalized to \([0,1]\). 
Table~\ref{tab:dataset_metrics} summarizes the evaluation protocol.

\begin{table}[h]
\centering
\small
\caption{Datasets and evaluation protocol.}
\label{tab:dataset_metrics}
\begin{tabular}{lll}
\toprule
Dataset & Setting & Metric \\
\midrule
ScienceQA~\citep{lu2022scienceqa} & IID: train on ScienceQA, test on ScienceQA & Accuracy \\
MathVision~\citep{wang2024mathvision} & Transfer: train on ViRL39K, test on MathVision & Accuracy \\
DynaMath~\citep{zou2025dynamath} & Transfer: train on ViRL39K, test on DynaMath & Accuracy \\
MMMU-Pro~\citep{yue2025mmmupro} & Transfer: train on ViRL39K, test on MMMU-Pro & Accuracy \\
\bottomrule
\end{tabular}
\end{table}

\subsection{Base Model, Prompts, and Baselines}
\label{app:baseline_details}

All internal methods use the same Qwen3-VL base family and the same evaluation prompts. 
The fast reference policy is the Qwen3-VL direct-answer / instruct policy, and the slow reference policy is the Qwen3-VL thinking policy. 
The trainable policy is initialized from the same Qwen3-VL checkpoint family for all internal training methods. 
We use identical input prompts, image preprocessing, answer extraction, and evaluation scripts across Qwen3-Instruct, Qwen3-Thinking, Mix SFT, Reward SFT, single-reference BOLT, and \textsc{DR-BOLT}. 
Thus, matched trainable internal comparisons help isolate the training objective from prompt, model-family, and teacher-bank changes. The frozen fast/slow reference rows provide context rather than an objective-only comparison.

\paragraph{External baselines.}
We reproduce the external adaptive-thinking methods under the common data and evaluation protocol; reported score/length values are from our runs rather than copied from their papers. Appendix~\ref{app:pinned_configuration} records the implementation provenance. 
DualMindVLM learns controllable fast and slow thinking modes with mode supervision and GRPO~\citep{lin2025dualmindvlm}. 
FAST-GRPO uses difficulty-aware rewards and adaptive length regularization~\citep{xiao2025fast}. 
TON trains VLMs to decide whether to reason using thought dropout and reinforcement learning~\citep{wang2025ton}. 
We also include a reproduced GRPO baseline under our evaluation setup. 
Because external methods may differ in base model and training implementation, they are treated as reference comparisons; the most controlled comparisons are the matched internal baselines.

\paragraph{Internal baselines.}
\textbf{Qwen3-Instruct} is the fast direct-answer baseline. 
\textbf{Qwen3-Thinking} is the slow reasoning baseline. 
\textbf{Mix SFT} trains on the union of fast and slow responses with uniform weights. 
\textbf{Reward SFT} uses reward-dependent weights without BOLT normalization or behavior-reference correction. 
\textbf{Single-ref Fast} applies single-reference BOLT using the fast reference. 
\textbf{Single-ref Slow} applies single-reference BOLT using the slow reference. 
\textbf{\textsc{DR-BOLT}} uses the dual-reference objective with mode-wise correction.

\subsection{Response Bank Construction}
\label{app:response_bank}

For each input, we build an offline response bank containing fast and slow candidates. 
Unless otherwise specified, we use \(N_{\mathrm{f}}=7\) fast candidates and \(N_{\mathrm{s}}=3\) slow candidates per input, corresponding to an empirical fast sampling ratio \(q=0.7\). 
The desired reference prior is fixed to \(\alpha=0.5\). 
Thus, the main experiments use a non-balanced response bank, which makes the behavior-reference correction active:
\[
    \kappa(y_{\mathrm{f}})=\frac{\alpha}{q}=\frac{0.5}{0.7}=0.714,
    \quad
    \kappa(y_{\mathrm{s}})=\frac{1-\alpha}{1-q}=\frac{0.5}{0.3}=1.667.
\]
For exact correction experiments, we compute
\[
    \kappa_{\mathrm{exact}}(x,y)
    =
    \frac{
        \alpha\piF(y|x)+(1-\alpha)\piS(y|x)
    }{
        q\piF(y|x)+(1-q)\piS(y|x)
    }
\]
in log space using teacher-forced likelihoods under both references. 
For the default \textsc{DR-BOLT} run, we retain the known generation source and use the mode-wise correction in Proposition~\ref{prop:modewise}.

\subsection{Reward, Decoding, and Mode Detection}
\label{app:reward_decoding}

The reward is
\[
    r(x,y)=r_{\mathrm{acc}}(x,y)+\gamma r_{\mathrm{fmt}}(y)-\lambda \ell(y).
\]
Here \(r_{\mathrm{acc}}\) is the task correctness reward, \(r_{\mathrm{fmt}}\) checks valid answer or reasoning format, and \(\ell(y)\) is the normalized response length. 
We use the same reward implementation for all BOLT-style methods. 
Table~\ref{tab:reward_hparams} summarizes the reward and correction hyperparameters.

\begin{table}[h]
\centering
\small
\caption{Reward and correction hyperparameters used in the main experiments.}
\label{tab:reward_hparams}
\resizebox{0.9\textwidth}{!}{%
\begin{tabular}{ll}
\toprule
Hyperparameter & Value / Definition \\
\midrule
Target fast prior \(\alpha\) & \(0.5\) \\
Empirical fast sampling ratio \(q\) & \(0.7\) \\
Fast candidates \(N_{\mathrm{f}}\) & \(7\) per input \\
Slow candidates \(N_{\mathrm{s}}\) & \(3\) per input \\
BOLT temperature \(\beta\) & \(1.0\) \\
Format reward weight \(\gamma\) & \(0.1\) \\
Length penalty \(\lambda\) & \(0.05\) \\
Length cost \(\ell(y)\) & word length normalized by the maximum response length in the candidate group \\
Mode-wise \(\kappa(y_{\mathrm{f}})\) & \(0.714\) \\
Mode-wise \(\kappa(y_{\mathrm{s}})\) & \(1.667\) \\
\bottomrule
\end{tabular}%
}
\end{table}

For response generation and routing calibration, we use temperature \(T=0.7\). 
For repeated auto-mode decoding in Figure~\ref{fig:partition_routing}, each prompt is sampled \(20\) times. 
A generated response is classified as slow if it contains the reasoning marker \texttt{<think>} or follows the slow-thinking format; otherwise it is classified as fast. 
In the mixed benchmark tables, the slow-ratio row aggregates responses only across MathVision, DynaMath, and MMMU-Pro from the ViRL39K-trained checkpoint. The ScienceQA-trained checkpoint is separate. The 100-prompt repeated-decoding subset used for calibration and sensitivity is a distinct measurement population. 
For benchmark scoring, final answers are extracted with the same answer parser for all internal methods.

\subsection{Optimization Details}
\label{app:optimization_details}

All trainable models are optimized with PyTorch using weighted next-token likelihood. 
We use LoRA fine-tuning with AdamW. 
The optimization objective is the weighted SFT loss defined in Eq.~\eqref{eq:dr_bolt_loss}. 
We mask prompt tokens in the label sequence so that loss is computed only on response tokens. 
The same optimizer, response bank, reward values, and data order are used across internal baselines whenever applicable. 
Table~\ref{tab:training_hparams} summarizes the training configuration.

\begin{table}[h]
\centering
\small
\caption{Training configuration for reproduced internal experiments.}
\label{tab:training_hparams}
\resizebox{0.9\textwidth}{!}{%
\begin{tabular}{ll}
\toprule
Item & Value \\
\midrule
Framework & PyTorch \\
Fine-tuning method & LoRA weighted SFT \\
Optimizer & AdamW \\
Training steps & \(500\) for white-box optimization diagnostics; full benchmark runs use the same budget across matched baselines \\
Per-device batch size & \(1\) \\
Gradient accumulation & \(8\) \\
Number of GPUs & \(4\) \\
Effective batch size & \(32\) sequences \\
Prompt loss masking & Prompt tokens are masked with label value \(-100\) \\
Phase 0 warm-up & \(100\) optimizer steps, unweighted 1:1 fast/slow mixture \\
\bottomrule
\end{tabular}%
}
\end{table}

\subsection{Length Measurement}
\label{app:length_details}

The main paper reports response \emph{Length} measured by word count. 
We use word count rather than tokenizer-specific token count because compared systems may use different tokenizers or model variants. 
Length is computed on generated responses only, excluding the input prompt. 
Reasoning markers such as \texttt{<think>} and \texttt{</think>} are included because they appear in the generated output and contribute to inference cost. 
Tokenizer-specific token counts should only be compared within the same tokenizer and are not used as the main cross-model metric.

\subsection{Compute and Resource Measurement}
\label{app:resource_measurement}

All reproduced resource measurements are conducted on a machine with \(4\) NVIDIA A100 GPUs with \(80\)GB memory each and \(32\) CPU cores. 
All implementations are based on PyTorch. 
We report wall-clock training time in hours, peak GPU memory in GB, and speedup relative to the reproduced GRPO baseline. 
The reported time includes candidate generation, reward computation, weight construction, and policy optimization. 
For GRPO-style baselines, the time includes online rollout and policy update. 
For \textsc{DR-BOLT}, the time includes offline candidate generation and weighted supervised optimization. 
Peak memory is measured as the maximum observed per-GPU memory during training. 
Speedup is computed as
\[
    \mathrm{Speedup}(\mathrm{method})
    =
    \frac{
        \mathrm{Time}(\mathrm{GRPO})
    }{
        \mathrm{Time}(\mathrm{method})
    }.
\]
Thus, values larger than \(1\) indicate faster training. 
In our reproduced setup, GRPO takes \(202.4\) hours with \(70.2\)GB peak memory, while \textsc{DR-BOLT} takes \(69.7\) hours with \(60.2\)GB peak memory, corresponding to a \(2.90\times\) speedup and lower memory usage. 
The resource advantage comes from decoupling response generation from policy optimization, whereas GRPO-style methods couple online rollout and update.

\section{Additional Ablations}
\label{app:additional_ablations}

\subsection{Downstream Ablation of Behavior-Reference Correction}
\label{app:kappa_downstream}

The white-box analyses in Appendix~\ref{app:dual_ref_validation} show that the mode-wise correction closely approximates the exact behavior-reference correction. 
We further test whether this approximation preserves downstream task performance. 
This ablation must use \(\alpha\neq q\); otherwise \(\rho=\mu\) and \(\kappa(x,y)=1\), making correction variants indistinguishable. 
We therefore set \(\alpha=0.5,q=0.7\), where fast samples are over-sampled and slow samples require up-weighting. 
All variants use the same ViRL39K response bank, reward, optimizer, Phase 0 warm-up, and training budget.

\begin{table}[h]
\centering
\small
\caption{
Downstream ablation of behavior-reference correction under \(\alpha=0.5,q=0.7\). 
Each benchmark cell reports \(\text{Score}/\text{Length}\). 
Slow Ratio is reported in percentage points.
}
\label{tab:kappa_downstream_ablation}
\begin{tabular}{lcccc}
\toprule
Correction 
& MathVision 
& DynaMath 
& MMMU-Pro 
& Slow Ratio \\
\midrule
No \(\kappa\) 
& \(0.260 / 518.2\)  
& \(0.626 / 308.1\)  
& \(0.314 / 329.9\)  
& \(0.13\)  \\
Exact \(\kappa\) 
& \(0.623 / 1023.2\)  
& \(0.804 / 325.7\)  
& \(0.594 / 382.0\)  
& \(22.0\) \\
Mode-wise \(\kappa\) 
& \(0.606 / 1058.1\)  
& \(0.799 / 370.6\)  
& \(0.581 / 385.7\)  
& \(24.1\) \\
\bottomrule
\end{tabular}
\end{table}

The results show that behavior-reference correction is necessary in the non-balanced setting. 
Without \(\kappa\), the model is strongly biased toward the over-sampled fast mode: the slow ratio drops to \(0.13\%\), and transfer scores fall sharply, especially on MathVision and MMMU-Pro. 
Exact correction restores the contribution of slow samples and gives the strongest scores. 
Mode-wise correction closely matches exact correction: the score gaps are \(0.017\), \(0.005\), and \(0.013\) on MathVision, DynaMath, and MMMU-Pro, respectively. 
The response Lengths and slow ratios are also close, showing that the practical mode-wise approximation preserves the downstream behavior of exact correction while avoiding expensive dual-reference likelihood evaluation.

\section{Controlled Offline Baselines and Further Experimental Details}
\label{app:rebuttal_experiments}

\subsection{Comparison with Alternative Offline Objectives}
\label{app:offline_alternatives}

We compare \textsc{DR-BOLT} with multi-reference DPO (MR-DPO) and mode-conditioned reward-weighted SFT (Cond-RW). These methods use the same fast/slow response bank, scalar reward and length penalty, frozen references, Qwen3-VL-8B-Thinking initialization, and Phase-0 warm-up. Optimization and response-token budgets are matched. ScienceQA results use a ScienceQA-trained checkpoint; MathVision, DynaMath, and MMMU-Pro results use a ViRL39K-trained checkpoint. The cross-benchmark slow ratio aggregates the latter three evaluations only.

\paragraph{Multi-reference DPO.}
For each input, we construct preferences from strictly reward-ordered pairs of bank candidates, discard reward ties, and subsample pairs to match the response-token training budget. The reference is the exact mixture
\[
\rho(y|x)=\alpha\piF(y|x)+(1-\alpha)\piS(y|x),
\]
computed in log space from the two frozen reference likelihoods. The loss is
\begin{equation}
\mathcal{L}_{\mathrm{MR\text{-}DPO}}=-\E_{(x,y^+,y^-)}\log\sigma\!\left(
\beta_{\mathrm{DPO}}\left[
\log\frac{\pi_\theta(y^+|x)}{\rho(y^+|x)}-
\log\frac{\pi_\theta(y^-|x)}{\rho(y^-|x)}\right]\right),
\label{eq:mr_dpo_baseline}
\end{equation}
with \(\beta_{\mathrm{DPO}}=0.1\). Each optimizer step consumes an effective batch of 32 response sequences over four GPUs, with the same seed, LoRA configuration, and optimizer settings as \textsc{DR-BOLT}. This baseline tests mixture-reference preference optimization while converting scalar rewards to pairwise orderings.

\paragraph{Mode-conditioned reward-weighted SFT.}
Each candidate is prepended with its known source token \(z\in\{\texttt{<FAST>},\texttt{<SLOW>}\}\). With \(\Y(x)\) denoting the candidate bank, this baseline uses
\begin{equation}
\begin{aligned}
w_{\mathrm{Cond\text{-}RW}}(x,y)&=\frac{\exp(r(x,y)/\beta)}{\sum_{y'\in\Y(x)}\exp(r(x,y')/\beta)},\\
\mathcal{L}_{\mathrm{Cond\text{-}RW}}&=-\sum_{(x,z,y)}w_{\mathrm{Cond\text{-}RW}}(x,y)\log\pi_\theta(z,y|x).
\end{aligned}
\label{eq:conditioned_rw_baseline}
\end{equation}
At evaluation time, no mode token is supplied externally: the model first generates the token and then the response. Thus the comparison uses one adaptive generation model without oracle mode information. Unlike \textsc{DR-BOLT}, it has no behavior-reference correction.

MR-DPO is the stronger alternative in Table~\ref{tab:offline_alternatives}, but \textsc{DR-BOLT} obtains higher scores and shorter responses on all four benchmarks in this setting. The comparison controls for the stronger slow teacher and candidate coverage. It supports the observed benefit of the full objective for this fixed scalar-reward bank; it does not establish universal superiority over pairwise preference learning, or isolate the causal contribution of each objective component individually.

\subsection{Hyperparameter Sensitivity}
\label{app:sensitivity}

We vary one factor at a time around \(\alpha=0.5,q=0.7,\beta=1.0,\lambda=0.05\), keeping the remaining reward, model, decoding, and optimization settings fixed. The target prior \(\alpha\) expresses a desired fast/slow preference, \(q\) controls candidate collection, \(\beta\) controls reward-tilting sharpness, and \(\lambda\) controls the accuracy--length preference. For the \(q\) sweep, equal-size ten-candidate banks are drawn from a common fast/slow candidate superset, so only their mode proportions change.

\begin{table}[t]
\centering
\small
\caption{One-factor ScienceQA sensitivity study. Length is mean response words. Slow ratio is measured on the 100-prompt repeated-decoding calibration subset used for Figure~\ref{fig:partition_routing}, not the full ScienceQA test set.}
\label{tab:scienceqa_sensitivity}
\begin{tabular}{lccc}
\toprule
Change from default & Score & Length & Calibration slow ratio (\%) \\
\midrule
Default & \(0.967\) & \(20.4\) & \(51.4\) \\
\(\alpha=0.3\) & \(0.968\) & \(25.5\) & \(71.1\) \\
\(\alpha=0.7\) & \(0.964\) & \(15.2\) & \(31.2\) \\
\(q=0.3\) & \(0.965\) & \(21.4\) & \(53.8\) \\
\(q=0.5\) & \(0.966\) & \(20.8\) & \(52.2\) \\
\(\beta=0.5\) & \(0.966\) & \(17.8\) & \(45.1\) \\
\(\beta=2.0\) & \(0.963\) & \(23.7\) & \(57.8\) \\
\(\lambda=0\) & \(0.968\) & \(28.9\) & \(59.2\) \\
\(\lambda=0.025\) & \(0.969\) & \(24.6\) & \(55.4\) \\
\(\lambda=0.10\) & \(0.964\) & \(15.7\) & \(43.8\) \\
\bottomrule
\end{tabular}
\end{table}

The prior \(\alpha\) changes slow usage monotonically: \(\alpha=0.3\) and \(0.7\) yield calibration slow ratios of \(71.1\%\) and \(31.2\%\). After correction, changing \(q\) from \(0.7\) to \(0.3\) changes score by \(0.002\) and calibration slow ratio by \(2.4\) percentage points. This is consistent with removing the systematic preference introduced by candidate collection, while finite-bank variance can remain. The \(\lambda\) sweep produces shorter responses as cost preference increases; the reported scores lie in \([0.964,0.969]\). Slow labels use the generated reasoning format, however short the trace, so a high slow-format fraction can coexist with a short mean response length.

We also train ViRL39K checkpoints for the same \(\lambda\) sweep and evaluate each checkpoint on all three transfer benchmarks without selecting \(\lambda\) separately on their test sets. Table~\ref{tab:transfer_length_sensitivity} reports MathVision and MMMU-Pro; its slow ratio aggregates all three benchmarks.

\begin{table}[t]
\centering
\small
\caption{Shared length-penalty sweep after ViRL39K training. Cells report score / mean response words; slow ratio aggregates MathVision, DynaMath, and MMMU-Pro.}
\label{tab:transfer_length_sensitivity}
\begin{tabular}{cccc}
\toprule
\(\lambda\) & MathVision & MMMU-Pro & Cross-benchmark slow ratio (\%) \\
\midrule
\(0\) & \(0.609/1396.3\) & \(0.578/512.8\) & \(31.8\) \\
\(0.025\) & \(0.614/1208.6\) & \(0.585/445.2\) & \(27.7\) \\
\(0.05\) & \(0.606/1058.1\) & \(0.581/385.7\) & \(24.1\) \\
\(0.10\) & \(0.589/811.4\) & \(0.561/302.7\) & \(17.4\) \\
\bottomrule
\end{tabular}
\end{table}

Both benchmarks become shorter as \(\lambda\) increases, with accuracy degrading under stronger compression. Their highest reported scores in this sweep occur at \(\lambda=0.025\); the shared main value \(0.05\) gives shorter responses with scores of \(0.606\) and \(0.581\). Thus the results characterize a utility trade-off rather than an accuracy-maximizing choice of \(\lambda\). Matched reward-based internal methods and both added offline baselines share the same reward and penalty. External adaptive-thinking reproductions preserve method-specific reward formulations and are therefore method-level comparisons rather than strictly \(\lambda\)-controlled ones.

\subsection{Learned Routers and Deployment Controls}
\label{app:learned_router}

The two-policy learned router retains the frozen Qwen3-VL-8B-Instruct and Qwen3-VL-8B-Thinking experts. For each training input, a lightweight classifier on frozen multimodal representations predicts
\[
z^*(x)=\arg\max_{z\in\{\mathrm{f},\mathrm{s}\}}U_z(x),\quad
U_{\mathrm{f}}(x)=\log\alpha+\log\widehat Z_{\mathrm{f}}(x),\quad
U_{\mathrm{s}}(x)=\log(1-\alpha)+\log\widehat Z_{\mathrm{s}}(x).
\]
The classifier chooses one expert at inference. No evaluation-set correctness labels or rewards are used to train or operate it. A separate one-checkpoint prompt-router diagnostic uses the same classifier to choose direct-answer or reasoning prompts for the unadapted thinking checkpoint. It tests whether selecting a prompt is sufficient to realize both behaviors within an existing model.

\begin{table}[t]
\centering
\small
\caption{Router controls. ScienceQA and transfer results use separately trained systems. Cells report score / mean response words; slow ratio aggregates the three transfer benchmarks.}
\label{tab:router_controls}
\resizebox{\linewidth}{!}{%
\begin{tabular}{lccccc}
\toprule
Method & ScienceQA & MathVision & DynaMath & MMMU-Pro & Slow ratio (\%) \\
\midrule
Two-policy router & \(0.958/64.7\) & \(0.401/2018.5\) & \(0.681/827.8\) & \(0.416/1054.6\) & \(28.9\) \\
Thinking prompt router & \(0.953/252.4\) & \(0.366/5108.7\) & \(0.652/2396.4\) & \(0.375/2748.9\) & \(87.6\) \\
\textsc{DR-BOLT} & \(0.967/20.4\) & \(0.606/1058.1\) & \(0.799/370.6\) & \(0.581/385.7\) & \(24.1\) \\
\bottomrule
\end{tabular}%
}
\end{table}

For transfer evaluation, the router classifier is trained on ViRL39K and the expert checkpoints remain frozen; \textsc{DR-BOLT} trains the generation policy itself on ViRL39K. The prompt router has \(18.4\%\) fast-prompt compliance, defined as the fraction of inputs routed to the direct-answer prompt producing neither a reasoning marker nor more than 30 generated tokens. Its \(87.6\%\) slow-format rate shows that selecting a direct-answer prompt does not reliably induce fast behavior in the thinking checkpoint.

The two-policy router needs two generation checkpoints and a separate classifier, whereas \textsc{DR-BOLT} needs one generation policy and no external router. This is a deployment distinction; we do not infer an exact serving-memory multiplier or latency advantage without measuring the loading strategy. The router comparison also changes whether generation policies are frozen or fine-tuned, so it does not isolate routing removal alone.

\subsection{Deterministic Reward Extraction}
\label{app:answer_extraction}

The training reward is \(r=r_{\mathrm{acc}}+0.1r_{\mathrm{fmt}}-0.05\ell(y)\), where correctness and format validity are binary and \(\ell(y)\) is response word count divided by the longest response word count in the same input's candidate group. Correctness uses the extracted final answer; intermediate reasoning is not graded. The format term checks for a parseable final answer in the required format rather than semantic correctness or plausible reasoning. Parse failures receive zero correctness and zero format reward. A format bonus of at most \(0.1\) and length penalty of at most \(0.05\) cannot make an incorrect answer outrank a correctly parsed correct answer solely through formatting. All internal methods share the same parsers and ground-truth annotations; no LLM judge is used for rewards or evaluation.

\paragraph{Answer span and option extraction.}
The parser uses the text after the last \texttt{</think>} marker, or the response when there is no closing marker. ScienceQA and the multiple-choice MMMU-Pro protocol upper-case the extracted letter, require it to fall within the item's valid option range, and compare it exactly with the ground-truth option letter. MMMU-Pro uses the item's supplied options, normally A--J in the ten-option setting. ScienceQA ground-truth letters are derived from the official answer index; answer-text matching is not used. The disclosed letter-extraction precedence is:
\begin{enumerate}
\item the last \texttt{\textbackslash boxed\{...\}} whose first character is a valid option letter;
\item a leading \texttt{A.} or \texttt{(A)} pattern;
\item an \texttt{answer is}, \texttt{answer:}, \texttt{answer =}, or \texttt{answer should be} cue followed by an optionally bolded letter;
\item a leading \texttt{**A**};
\item a span consisting of exactly one letter;
\item a leading letter followed by whitespace;
\item any line reducing to a single letter after stripping surrounding periods and parentheses.
\end{enumerate}
Unrecoverable or out-of-range letters are incorrect. The parser also rejects a response naming multiple options, such as \texttt{E, H}, as a parse failure.

\paragraph{MathVision and DynaMath normalization.}
For MathVision items with options and a single-letter ground truth, the letter path uses the item's option range; the disclosed normalization additionally accepts the ground-truth option surface text as a case-insensitive substring. Otherwise, the numeric path uses the last boxed span when present, or the answer span, and takes its last numeric literal. The disclosed numeric pattern is
\begin{center}
\texttt{-?\textbackslash d+(\textbackslash .\textbackslash d+)?([eE][+-]?\textbackslash d+)?}
\end{center}
A parsed numeric value \(v\) matches ground truth \(g\) when \(|v-g|\leq 10^{-3}\max(1,|g|)\). For non-numeric ground truth, the disclosed fallback uses case-insensitive substring containment. DynaMath uses the same numeric extraction and tolerance, or substring matching for short non-numeric answers. Unparseable responses are incorrect. These are deterministic normalization rules and should not be interpreted as a general symbolic-equivalence verifier.

The implementation uses the MathVision evaluator's extraction and equality utilities, the DynaMath scoring routine (an accuracy sub-score above \(0.5\) is treated as correct), and the MMMU multiple-choice parser for MMMU-Pro.

\begin{table}[t]
\centering
\small
\caption{Representative parsing cases from generated responses.}
\label{tab:parser_examples}
\begin{tabular}{@{}p{0.17\linewidth}p{0.38\linewidth}p{0.37\linewidth}@{}}
\toprule
Dataset & Answer span & Parsing outcome \\
\midrule
ScienceQA & \texttt{B. New Hampshire} & Leading-option rule; B matches ground truth B. \\
ScienceQA & \texttt{B} after a 168-word slow response & Reasoning is discarded; single-letter rule yields B. \\
ScienceQA & \texttt{Neither.} & No option is recovered; zero correctness and format reward. \\
MathVision & Unboxed arithmetic-sequence derivation & Last numeric literal is 60; matches ground truth 60. \\
MathVision & \texttt{A} for a multiple-choice item & Letter path yields A; matches the correct option. \\
MMMU-Pro & \texttt{E, H} & Multiple options; no unique answer, so zero correctness and format reward. \\
\bottomrule
\end{tabular}
\end{table}

\subsection{Pinned Checkpoints, Optimization, and Cost Provenance}
\label{app:pinned_configuration}

The exact checkpoints and snapshot revisions used for internal runs are listed in Table~\ref{tab:checkpoint_revisions}. The processor and tokenizer use the corresponding model snapshot. The FAST preview checkpoint is used to verify the Qwen3-VL-8B setup; the dual-reference candidates come from the Instruct and Thinking checkpoints.

\begin{table}[t]
\centering
\small
\caption{Model snapshot revisions.}
\label{tab:checkpoint_revisions}
\begin{tabular}{p{0.30\linewidth}p{0.62\linewidth}}
\toprule
Role & Checkpoint and revision \\
\midrule
Fast reference & \texttt{Qwen/Qwen3-VL-8B-Instruct}\newline\texttt{0c351dd01ed87e9c1b53cbc748cba10e6187ff3b} \\
Slow reference and initialization & \texttt{Qwen/Qwen3-VL-8B-Thinking}\newline\texttt{92f3c4b4feadd3a016ef468d103bb5f58b2a2c6b} \\
FAST setup verification & \texttt{xiaowenyi/FAST-8B-Preview}\newline\texttt{45aa28a69976f14f295f34d4bde25099a84cf1e5} \\
\bottomrule
\end{tabular}
\end{table}

Phase 0 is 100 optimizer steps of unweighted SFT on a 1:1 fast/slow mixture. It uses the same LoRA modules, AdamW optimizer, learning rate, and batch configuration as Phase 2, without reward or behavior-reference weights. The shared configuration is summarized in Table~\ref{tab:extended_training_configuration}.

\begin{table}[t]
\centering
\small
\caption{Additional shared optimization and decoding configuration.}
\label{tab:extended_training_configuration}
\begin{tabular}{p{0.23\linewidth}p{0.69\linewidth}}
\toprule
Item & Value \\
\midrule
LoRA & Rank 8, scaling factor 64, dropout 0.05, no bias. \\
Target modules & \texttt{q\_proj}, \texttt{k\_proj}, \texttt{v\_proj}, \texttt{o\_proj}, \texttt{gate\_proj}, \texttt{up\_proj}, \texttt{down\_proj}. \\
AdamW & Learning rate \(5\times10^{-6}\), weight decay \(0.1\), moments \((0.9,0.95)\). \\
Schedule & Cosine decay with warm-up ratio \(0.1\). \\
Batch & Four GPUs, per-device batch 1, gradient accumulation 8; effective batch 32 response sequences. \\
Precision / attention & bf16 / SDPA; key-value cache disabled during training. \\
Masking & Prompt tokens and assistant header have label \(-100\). \\
Checkpoint cadence & Every 25 optimizer steps for implicit-KL replay. \\
Decoding & Temperature \(0.7\), top-p \(0.9\); maximum new tokens 2048 for ScienceQA and 8192 for ViRL39K, MathVision, DynaMath, and MMMU-Pro. \\
Diagnostics & 500 steps; 200 held-out samples replayed every 25 steps. \\
\bottomrule
\end{tabular}
\end{table}

The random seed for LoRA initialization and data order is shared across matched buckets, and the full benchmark optimization budget is matched. The 500-step budget refers to the white-box diagnostic experiment.

All reported external score/length entries are our common-data reproductions, rather than scores copied from their original papers. FAST-GRPO follows its released recipe. TON retains its thought-dropout SFT plus GRPO procedure with ViRL39K converted to its expected multimodal schema. DualMindVLM is reimplemented from its two-stage mode-labeling and GRPO specification. Their cross-benchmark training data are ViRL39K and their evaluation inputs and answer parser are shared. The matched internal suite additionally controls references, candidates, reward values, initialization, optimizer, and data order wherever applicable.

\begin{table}[t]
\centering
\small
\caption{Runtime provenance on four A100-80GB GPUs. Times are wall-clock hours from our reproduced runs.}
\label{tab:runtime_provenance}
\begin{tabular}{lcp{0.61\linewidth}}
\toprule
Method & Time (h) & Included stages \\
\midrule
GRPO & \(202.4\) & Online rollout generation and policy updates. \\
FAST-GRPO & \(176.6\) & Online rollout generation and policy updates under the released FAST recipe. \\
TON & \(248.9\) & Thought-dropout SFT, online rollouts, and policy updates. \\
DualMindVLM & -- & Runtime is not reported. \\
\textsc{DR-BOLT} & \(69.7\) & Fast/slow candidate generation, correctness and format rewards, weight construction, and policy optimization. \\
\bottomrule
\end{tabular}
\end{table}

The \(2.90\times\) speedup is \(202.4/69.7\) and therefore includes offline response generation. Peak per-GPU memory is \(60.2\)GB for \textsc{DR-BOLT} and \(70.2\)GB for GRPO. The response bank, references, reward values, and weights are frozen before Phase-2 optimization, so no new on-policy rollouts or reference-likelihood recomputation are needed after each update.

\subsection{Candidate Coverage and Scope of the Evidence}
\label{app:candidate_coverage}

The union of seven fast and three slow candidates on ViRL39K has reported oracle coverage
\[
A_{\mathrm{bank}}=\Pr\!\left[\exists y\in\Y_{\mathrm{f}}(x)\cup\Y_{\mathrm{s}}(x):r_{\mathrm{acc}}(x,y)=1\right]\approx0.947.
\]
This is the fraction of training inputs with at least one correct candidate, not a deployed model score or teacher pass@1 accuracy. Reward tilting can exploit complementary candidates, so either reference's single-decoding accuracy is not a strict ceiling on the learned policy. However, the finite bank supplies reward evidence only for sampled candidates and does not provide unconstrained online exploration.

\subsection{Additional Same-Benchmark Evaluation on 3DSRBench}
\label{app:3dsrbench}

We additionally evaluate on 3DSRBench~\citep{ma20253dsrbench}, which covers height, location, orientation, viewpoint, and multi-object 3D spatial reasoning. Its real, synthetic-common, and synthetic-uncommon subsets are evaluation groups rather than a supervised train/test split. We therefore construct a disjoint training/test partition. The response bank and \textsc{DR-BOLT} training use only the training partition; all three methods in Table~\ref{tab:3dsrbench} are evaluated on the identical held-out test partition. No test examples or test rewards enter candidate generation or training. This is a same-benchmark experiment, separate from ViRL39K transfer evaluation.

\begin{table}[t]
\centering
\small
\caption{3DSRBench results on the constructed held-out test partition. Length is mean response words.}
\label{tab:3dsrbench}
\begin{tabular}{lcc}
\toprule
Method & Score & Length \\
\midrule
Fast reference & \(0.531\) & \(13.9\) \\
Slow reference & \(0.548\) & \(1846.1\) \\
\textsc{DR-BOLT} & \(0.546\) & \(452.2\) \\
\bottomrule
\end{tabular}
\end{table}

\textsc{DR-BOLT} retains nearly all of the slow reference's accuracy, with a difference of \(0.002\), while reducing mean length by \(75.5\%\) (\(4.08\times\)). It improves on the fast reference by \(0.015\). This illustrates that adaptive reasoning can retain most of the correctness benefit of slow reasoning while removing unnecessary length; the intended utility trade-off does not require exceeding the slow reference in raw accuracy.

\FloatBarrier

\section{Qualitative Examples}
\label{app:qual_examples_full}

We provide qualitative examples to illustrate how \textsc{DR-BOLT} allocates fast and slow thinking across different inputs. 
Case~1 is a ViRL39K training example comparing fast and slow teacher responses. 
Cases~2--4 are held-out transfer examples from MathVision, DynaMath, and MMMU-Pro. 
The examples show that \textsc{DR-BOLT} uses reasoning when visual pattern or chart interpretation is needed, while producing direct answers for visually obvious symbolic recognition questions.

\begin{tcolorbox}[
    colback=white,
    colframe=black,
    coltitle=white,
    colbacktitle=black,
    title=\textbf{Case 1: ViRL39K Training Example},
    fonttitle=\bfseries,
    arc=2mm,
    boxrule=0.8pt
]
\centering
\includegraphics[width=0.62\linewidth]{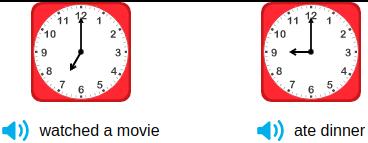}

\vspace{0.5em}
\raggedright
\textbf{Question:} The clocks show two things Lily did Tuesday evening. Which did Lily do earlier?  
Choices: (A) ate dinner; (B) watched a movie.  
Answer the question directly in one \(\backslash\)boxed\{\}.  

\textbf{Ground Truth:} \(\boxed{B}\)

\vspace{0.5em}
\textbf{Fast teacher output} \textit{(169 words):}  
The fast teacher reads the clocks as 7:00 for ``watched a movie'' and 9:00 for ``ate dinner'', compares the two times, and correctly answers \(\boxed{B}\).

\vspace{0.3em}
\textbf{Slow teacher output} \textit{(278 words):}  
The slow teacher gives the same correct answer, but produces a longer reasoning trace that repeats the clock-reading process.  
This case shows that for easy visual reasoning examples, slow reasoning is correct but unnecessary.
\end{tcolorbox}

\begin{tcolorbox}[
    colback=white,
    colframe=black,
    coltitle=white,
    colbacktitle=black,
    title=\textbf{Case 2: MathVision Transfer Example},
    fonttitle=\bfseries,
    arc=2mm,
    boxrule=0.8pt
]
\centering
\includegraphics[width=0.85\linewidth]{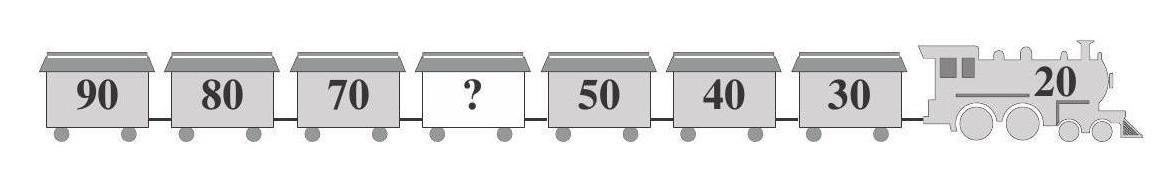}

\vspace{0.5em}
\raggedright
\textbf{Question:} Which number should be written in place of the question mark?

\textbf{Ground Truth:} \(60\)

\vspace{0.5em}
\textbf{\textsc{DR-BOLT} output} \textit{(180 tokens):}  
The model identifies the sequence in the train cars:
\[
90,\;80,\;70,\;?,\;50,\;40,\;30,\;20.
\]
It recognizes that the numbers decrease by \(10\) at each step, infers that the missing value is \(70-10=60\), and outputs \(60\).

\vspace{0.3em}
\textbf{Observation:}  
\textsc{DR-BOLT} activates a short reasoning process because solving the problem requires recognizing a visual numerical pattern.
\end{tcolorbox}

\begin{tcolorbox}[
    colback=white,
    colframe=black,
    coltitle=white,
    colbacktitle=black,
    title=\textbf{Case 3: DynaMath Transfer Example},
    fonttitle=\bfseries,
    arc=2mm,
    boxrule=0.8pt
]
\centering
\includegraphics[width=0.70\linewidth]{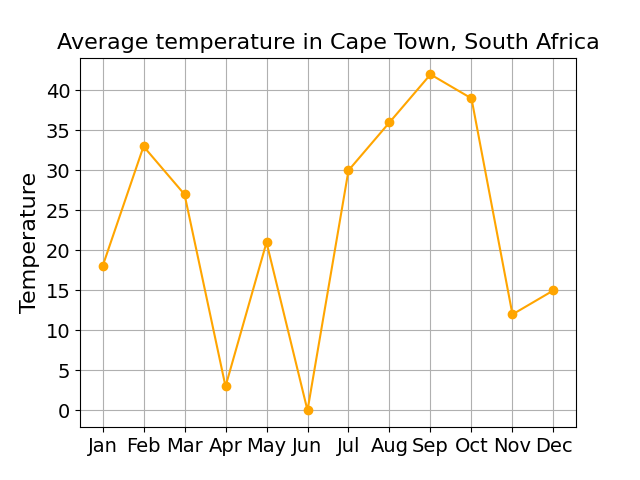}

\vspace{0.5em}
\raggedright
\textbf{Question:} Use the graph to answer the question below. Which month is the hottest on average in Cape Town?

\textbf{Ground Truth:} Sep

\vspace{0.5em}
\textbf{\textsc{DR-BOLT} output} \textit{(138 tokens):}  
The model inspects the line graph, compares the monthly temperature values, and identifies the highest point on the curve. 
The peak occurs in September, where the temperature is slightly above \(40^\circ\)C. 
It therefore answers September.

\vspace{0.3em}
\textbf{Observation:}  
The model uses reasoning to inspect the chart and compare values, rather than relying on a direct recognition shortcut.
\end{tcolorbox}

\begin{tcolorbox}[
    colback=white,
    colframe=black,
    coltitle=white,
    colbacktitle=black,
    title=\textbf{Case 4: MMMU-Pro Transfer Example},
    fonttitle=\bfseries,
    arc=2mm,
    boxrule=0.8pt
]
\centering
\includegraphics[width=0.42\linewidth]{fig/case4}

\vspace{0.5em}
\raggedright
\textbf{Question:} The image is a symbol of which option?  
Options include:  
A. A memory; B. A transition; C. An opportunity; D. A common experience; E. A journey; F. A warning; G. A clock; H. Time waits for no man; I. An obstacle; J. A mystery.  
Answer with the option letter only.

\textbf{Ground Truth:} H

\vspace{0.5em}
\textbf{\textsc{DR-BOLT} output} \textit{(2 tokens):}  
H

\vspace{0.3em}
\textbf{Observation:}  
The image directly depicts Father Time, which corresponds to the idiom ``Time waits for no man.'' 
\textsc{DR-BOLT} answers directly without unnecessary reasoning, showing that the model can preserve fast behavior on visually obvious symbolic questions.
\end{tcolorbox}

\FloatBarrier


\clearpage
\section*{NeurIPS Paper Checklist}

\begin{enumerate}

\item {\bf Claims}
    \item[] Question: Do the main claims made in the abstract and introduction accurately reflect the paper's contributions and scope?
    \item[] Answer: \answerYes{}.
    \item[] Justification: The abstract and introduction state the main contributions: dual-reference offline optimization, the DR-BOLT objective, router-free fast/slow switching, and empirical improvements on ScienceQA and OOD multimodal reasoning benchmarks. The claims are supported by theoretical derivations in Sections~\ref{sec:prelim}--\ref{sec:dr_bolt} and experiments in Section~\ref{sec:experiments}.

\item {\bf Limitations}
    \item[] Question: Does the paper discuss the limitations of the work performed by the authors?
    \item[] Answer: \answerYes{}.
    \item[] Justification: We include a Limitations section discussing the two-mode setting, the approximate mode-format separation assumption, and the fact that runtime may vary with hardware, inference engine, rollout length, and implementation details.

\item {\bf Theory assumptions and proofs}
    \item[] Question: For each theoretical result, does the paper provide the full set of assumptions and a complete (and correct) proof?
    \item[] Answer: \answerYes{}.
    \item[] Justification: The assumptions are stated in the main text, including the approximate mode-format separation assumption. Proofs for single-reference BOLT, dual-reference optimality, dual-reference weighted SFT, routing by partition competition, and mode-wise correction are provided in the appendix.

\item {\bf Experimental result reproducibility}
    \item[] Question: Does the paper fully disclose all the information needed to reproduce the main experimental results of the paper to the extent that it affects the main claims and/or conclusions of the paper (regardless of whether the code and data are provided or not)?
    \item[] Answer: \answerYes{}.
    \item[] Justification: The paper describes the datasets, IID/OOD split protocol, baselines, length measurement, resource measurement, and evaluation metrics. All datasets used in the paper are publicly available. We will release the code and model weights with instructions to reproduce the main results.

\item {\bf Open access to data and code}
    \item[] Question: Does the paper provide open access to the data and code, with sufficient instructions to faithfully reproduce the main experimental results, as described in supplemental material?
    \item[] Answer: \answerYes{}.
    \item[] Justification: The datasets are public benchmarks. Our code is available at https://github.com/HongbinLin3589/DR-BOLT.

\item {\bf Experimental setting/details}
    \item[] Question: Does the paper specify all the training and test details (e.g., data splits, hyperparameters, how they were chosen, type of optimizer) necessary to understand the results?
    \item[] Answer: \answerYes{}.
    \item[] Justification: The paper specifies that ScienceQA is used for IID training and evaluation, while ViRL39K is used for training and MathVision, DynaMath, and MMMU-Pro are used for OOD evaluation. The appendix describes response-bank construction, reward components, correction settings, length measurement, and resource measurement. Additional implementation details will be included in the released code.

\item {\bf Experiment statistical significance}
    \item[] Question: Does the paper report error bars suitably and correctly defined or other appropriate information about the statistical significance of the experiments?
    \item[] Answer: \answerNo{}.
    \item[] Justification: We report confidence intervals for the partition-based routing calibration and provide repeated-decoding statistics for routing behavior. However, due to the computational cost of training large VLMs, we do not report multi-seed error bars for all benchmark scores. We provide detailed ablations and controlled comparisons under the same hardware and data scale.

\item {\bf Experiments compute resources}
    \item[] Question: For each experiment, does the paper provide sufficient information on the computer resources (type of compute workers, memory, time of execution) needed to reproduce the experiments?
    \item[] Answer: \answerYes{}.
    \item[] Justification: The appendix reports the compute setting: 4 NVIDIA A100 80GB GPUs and 32 CPU cores. The main tables report wall-clock training time, peak GPU memory, and speedup relative to reproduced GRPO. The reported time includes candidate generation, reward computation, weight construction, and policy optimization.

\item {\bf Code of ethics}
    \item[] Question: Does the research conducted in the paper conform, in every respect, with the NeurIPS Code of Ethics \url{https://neurips.cc/public/EthicsGuidelines}?
    \item[] Answer: \answerYes{}.
    \item[] Justification: The work uses public datasets and standard model training/evaluation procedures. We are not aware of any deviation from the NeurIPS Code of Ethics.

\item {\bf Broader impacts}
    \item[] Question: Does the paper discuss both potential positive societal impacts and negative societal impacts of the work performed?
    \item[] Answer: \answerYes{}.
    \item[] Justification: The method may reduce unnecessary inference cost and energy use by avoiding overthinking on simple inputs. As with other VLM reasoning methods, incorrect model outputs or overconfidence may still cause downstream risks if deployed in high-stakes settings; our work focuses on research benchmarks rather than deployment.

\item {\bf Safeguards}
    \item[] Question: Does the paper describe safeguards that have been put in place for responsible release of data or models that have a high risk for misuse (e.g., pre-trained language models, image generators, or scraped datasets)?
    \item[] Answer: \answerNA{}.
    \item[] Justification: The paper does not introduce a new scraped dataset or a general-purpose foundation model from scratch. We fine-tune existing VLMs for research on adaptive reasoning using public benchmarks. Released weights will follow the license and usage terms of the base models.

\item {\bf Licenses for existing assets}
    \item[] Question: Are the creators or original owners of assets (e.g., code, data, models), used in the paper, properly credited and are the license and terms of use explicitly mentioned and properly respected?
    \item[] Answer: \answerYes{}.
    \item[] Justification: We cite all datasets and baseline methods used in the experiments, including ScienceQA, ViRL39K, MathVision, DynaMath, MMMU-Pro, Qwen models, and external adaptive-thinking baselines. We will document dataset and model licenses in the code/model release.

\item {\bf New assets}
    \item[] Question: Are new assets introduced in the paper well documented and is the documentation provided alongside the assets?
    \item[] Answer: \answerYes{}.
    \item[] Justification: We will release code and model weights as new research artifacts. The release will include documentation describing training data, evaluation scripts, model checkpoints, expected usage, and limitations.

\item {\bf Crowdsourcing and research with human subjects}
    \item[] Question: For crowdsourcing experiments and research with human subjects, does the paper include the full text of instructions given to participants and screenshots, if applicable, as well as details about compensation (if any)? 
    \item[] Answer: \answerNA{}.
    \item[] Justification: The paper does not involve crowdsourcing or human-subject experiments. All experiments use public datasets and automated evaluation protocols.

\item {\bf Institutional review board (IRB) approvals or equivalent for research with human subjects}
    \item[] Question: Does the paper describe potential risks incurred by study participants, whether such risks were disclosed to the subjects, and whether Institutional Review Board (IRB) approvals (or an equivalent approval/review based on the requirements of your country or institution) were obtained?
    \item[] Answer: \answerNA{}.
    \item[] Justification: The paper does not involve human-subject research, so IRB approval is not applicable.

\item {\bf Declaration of LLM usage}
    \item[] Question: Does the paper describe the usage of LLMs if it is an important, original, or non-standard component of the core methods in this research? Note that if the LLM is used only for writing, editing, or formatting purposes and does \emph{not} impact the core methodology, scientific rigor, or originality of the research, declaration is not required.
    \item[] Answer: \answerYes{}.
    \item[] Justification: VLMs are central to the method and experiments. The paper describes the fast and slow VLM policies, the trainable policy, and the use of public VLM checkpoints for response generation, training, and evaluation.

\end{enumerate}

\end{document}